\documentclass[10pt,journal,compsoc]{IEEEtran}
\usepackage[T1]{fontenc}
\usepackage[latin9]{inputenc}
\usepackage{color}
\usepackage{amsmath}
\usepackage{amssymb}
\usepackage{graphicx}

\makeatletter

\providecommand{\tabularnewline}{\\}

\ifCLASSOPTIONcompsoc
  \usepackage[nocompress]{cite}
\else
  \usepackage{cite}
\fi

\usepackage{xcolor}\usepackage{caption}
\usepackage{todonotes}

\DeclareMathOperator*{\argmax}{\arg\!\max}
\usepackage{algorithm}
\usepackage{algorithmic}
\usepackage{varwidth}

\providecommand{\tabularnewline}{\\}

\newcommand*{\pl}{ }
\newcommand*{\etal}{{\it et al.}}

\makeatother

\begin{document}
\title{Heterogeneous Knowledge Transfer in Video Emotion Recognition, Attribution and Summarization}
\author{Baohan~Xu, Yanwei~Fu, Yu-Gang~Jiang, Boyang~Li and Leonid~Sigal
\IEEEcompsocitemizethanks{ \IEEEcompsocthanksitem Baohan Xu and Yu-Gang Jiang are with the School of Computer Science, Shanghai Key Lab of Intelligent Information Processing, Fudan University, Shanghai, China. Email:\{bhxu14,ygj\}@fudan.edu.cn. \IEEEcompsocthanksitem Yanwei~Fu (corresponding author) is with the School of  Data Science, Fudan University, Shanghai, China. Email:\{yanweifu\}@fudan.edu.cn \IEEEcompsocthanksitem  Boyang Li and Leonid Sigal are with Disney Research. Email:  \{albert.li, lsigal\}@disneyresearch.com. } \thanks{} }

\IEEEtitleabstractindextext{ 
\begin{abstract} 
Emotion is a key element in user-generated videos. However, it is difficult to understand emotions conveyed in such videos due to the complex and unstructured nature of user-generated content and the sparsity of video frames expressing emotion. In this paper, for the first time, we study the problem of transferring knowledge from heterogeneous external sources, including image and textual data, to facilitate three related tasks in understanding video emotion: emotion recognition, emotion attribution and emotion-oriented summarization. Specifically, our framework (1) learns a video encoding from an auxiliary emotional image dataset in order to improve supervised video emotion recognition, and (2) transfers knowledge from an auxiliary textual corpora for zero-shot \pl{recognition} of emotion classes unseen during training. The proposed technique for knowledge transfer facilitates novel applications of emotion attribution and emotion-oriented summarization. A comprehensive set of experiments on multiple datasets demonstrate the effectiveness of our framework.
\end{abstract}
}
\maketitle

\section{Introduction\label{sec:introduction} }

Rapid development of mobile devices has led to an explosive growth
of user-generated images and videos, which creates a demand for computational
understanding of visual media content. In addition to recognition
of objective content, such as objects and scenes, an important dimension
of video content analysis is the understanding of emotional or affective
content, i.e. estimating the emotional impact of the video on a viewer.
Emotional content can strongly resonate with viewers and plays a crucial
role in the video-watching experience. Some successes have been achieved
with the use of deep-learning architectures trained for text at both
sentence- and document-level \cite{Kotzias2014DeepMultinstance} or
image sentiment analysis \cite{TaoChen2014Deepsentibank}. However,
the ability to understand emotions from video, to a large extent,
remains an unsolved problem.

Analysis of emotional content in video has many real-world applications.
Video recommendation services can benefit from matching user interests
with the emotions of video content and prediction of interestingness
\cite{yugangVideoInteresting2013,yanwei_interestingness,interestingnessECCV2014},
leading to improved user satisfaction. Better understanding of video
emotions may enable advertising that is consistent with the main video's
mood and help avoid social inappropriateness such as placing a funny
advertisement alongside a funeral video. Video summarization \cite{Truong:2007:VAS:1198302.1198305}
and coding \cite{Pan2015Efficient} can also benefit from understanding
emotions, since an accurate summary should keep the emotional content
conveyed by the original video.

Unlike professionally produced videos, user-generated video content
presents unique challenges for video understanding. Challenges arise
from the diversity of the content, lack of structure, and, typically,
poor production and editing quality (e.g., insufficient lighting).
Analyzing the video emotional content in such videos is even more
difficult, since (1) the complex spatio-temporal interactions between
visual elements makes this intrinsically more complex than analysis
of static images, and (2) the emotion is often expressed in only certain
limited (sparse) keyframes or video clips.

To cope with these difficulties we employ heterogeneous knowledge
extracted from external sources. In particular, we propose an auxiliary
Image Transfer Encoding (ITE) algorithm which can leverage emotional
information from auxiliary image data to aggregate frame-level features
into a video-level emotion-sensitive representation. To demonstrate
the power of this knowledge transfer technique, in this paper we tackle
three inter-related tasks, namely emotion recognition, emotion attribution,
and emotion-oriented summarization.

The first task of emotion recognition includes both supervised and
zero-shot conditions. Zero-shot video emotion recognition aims to
recognize emotion classes that are not seen during training. This
task is motivated by recent cognitive theories \cite{Barrett2006,Lindquist2012,Carroll1996,Barrett2007}
that suggest human emotional experiences extend beyond the traditional
``basic emotion'' categories (e.g., Ekman's six emotions \cite{Ekman1972}).
Rather, many cognitive processes cooperate closely to create rich
emotional and affective experiences \cite{Li:Humor2015,Marsella2009,Gross2002},
such as ecstasy, nostalgia, or suspense. When operating in the real
world, recognition systems trained with a small set of emotion labels
will inevitably encounter emotion types that are not present in its
training set. From large image and text corpora, we construct a semantic
vector space where we can identify semantic relationships between
visual representation and textual representation of emotions. Subsequently,
we can exploit the semantic relationships to recognize emotions unknown
to the system. To our best knowledge, our paper is the first to explore
zero-shot emotion recognition.

For our second task, we define a novel problem, \emph{video emotion
attribution}, which aims to identify each frame's contribution to
a video's overall emotion. This task is motivated by the observation
that emotional videos, even those conveying strong emotions, typically
contain many frames that are emotionally neutral. The neutral frames
may serve important functions, such as setting up the context,
but do not convey emotions themselves. Being able to detect emotional
frames is a key component of video understanding and enables our third
task.

Our third task is \emph{emotion-oriented video summarization}. We
argue that a good video summary should be succinct but also provide
good coverage of the original video's emotion and information content.
Hence our approach aims to balance emotion, information content, and
length in providing an accurate video summary.

\vspace{0.1in}
\noindent \textbf{Contributions:} We introduce a framework for transferring
knowledge from heterogeneous sources, including image and text, for
the understanding of emotions in videos. To the best of our knowledge,
this is the first work on zero-shot emotion recognition achieved by
applying knowledge learned from text sources to the video domain.
We also propose the first definition and solution for the problems
of emotion video attribution and emotion-oriented summarization. We
show that our emotion-oriented summaries are better than alternative
methods that do not consider emotion. Finally, we introduce, and will
make available to the community, two new emotion-centric video datasets:
VideoStory-P14 and YF-E6. \noindent 

\section{Related work}

\subsection{Psychological Theories of Emotion}

\noindent \label{section:psyc-theories} It is a widely held view
in psychology that emotions can be categorized into a number of static
categories, each of which is associated with stereotypical facial
expression(s), physiological measurements, behaviors, and external causes
\cite{Dolan2002}. The most well-known model is probably Ekman's six
pan-cultural basic emotions, including happiness, sadness, disgust,
anger, fear, and surprise \cite{Ekman1972,nature_emotion}. However,
the exact categories can vary from one model to another. Plutchik
\cite{Plutchik1980} added anticipation and trust to the list. Ortony,
Clore and Collins's \cite{OCC1988} model of emotion defined up to
22 emotions, including categories like hope, shame, and gratitude.

Nevertheless, more recent empirical findings and theories \cite{Barrett2006,Lindquist2012}
suggest emotional experiences are much more varied than previously
assumed. It has been argued that the classical categories are only
modal or stereotypical emotions, and large fuzzy areas exist in-between
on the emotional landscape. Other theories \cite{Li:Humor2015,Marsella2009,Gross2002}
highlight the dynamics of emotion and the interactions between emotional
processes and other cognitive processes. Together, the complex dynamics
and interactions produce a rich set of emotional and affective experiences,
and correspondingly rich natural language descriptions of those experiences,
such as ecstasy, nostalgia, or suspense.

In order to cope with diverse emotional descriptions that may be practically
difficult (or at least very costly) to label, in this paper, we investigate
emotion recognition in a zero-shot setting (in addition to, a traditional,
supervised setting). Our recognition system is tested against emotional
classes that do not appear in the training set. The zero-shot recognition
task puts to test the system's ability to effectively utilize knowledge
learned from heterogeneous sources in order to adapt to unseen emotional
labels.

This paper is mostly concerned with recognizing emotion aroused from
watching a video rather than recognizing facial expressions. Despite
inherent subjectivity involved in emotional experiences and individual
differences \cite{Hamann2004}, there are likely modal responses that
can be gathered from a reasonable and neutral audience. A number of
recent works focused on recognizing the emotional impact of images
and videos, as we review in the next section.

\subsection{Automatic Emotion Analysis}

In this section, we briefly review two relevant areas of research:
recognition of emotional impact of images on viewers, and recognition
of emotional impact from videos.

\vspace{0.1in}
\noindent \textbf{Recognizing emotional impact of still images on
viewers.} Machajdik and Hanbury \cite{Machajdik2010} classified images
into 8 affective categories: amusement, awe, contentment, excitement,
anger, disgust, fear, and sadness. In addition to color, texture,
and statistics about faces and skin area present in the image, they
also make use of composition features such as the rule of the third
and depth of field. Lu \etal \cite{Lu2012} studied shape features
along the dimensions of rounded-angular and simple-complex, and their
effects in arousing viewers' emotions. You \etal \cite{You2015AAAI_img_sentiment}
designed a deep convolutional neural network (CNN) for visual sentiment
analysis. After training on the entire training set, images on which
the CNN performs poorly are stochastically removed. The remaining
images were used to fine-tune the network. A few works \cite{TaoChen2014Deepsentibank,CanXu2014visual_senti_arxiv}
also employed off-the-shelf CNN features.

 \vspace{0.1in}
\noindent \textbf{Recognizing emotional impact of videos}. A Large
number of early works studied emotion in movies (e.g., \cite{Kang2003,Wang2006,Irie2010}).
Wang and Cheong \cite{Wang2006} used an SVM with diverse audio-visual
features to classify 2040 scenes in 36 Hollywood movies into 7 emotions.
Jou \emph{et al}. \cite{predictGIF2014ACMMM} worked on animated GIF
files. Irie \etal \cite{Irie2010} use Latent Dirichlet Allocation
to extract audio-visual topics as mid-level features, which are combined
with Hidden-Markov-like dynamic model. For a more comprehensive review
we refer reader to a survey \cite{Wang-Ji2015}.

SentiBank \cite{Borth2013acmmm} contains a set of $1,553$
adjective-noun pairs, such as ``beautiful flowers'' and ``sad eyes'',
and images exemplifying each pair. One linear SVM detector was trained
for each pair. The best-performing $1,200$ detectors provide a mid-level
representation for emotion recognition. Chen \emph{et al.} \cite{TaoChen2014Deepsentibank}
replaced the SVM detectors with deep convolutional neural networks.
Jiang \emph{et al.} \cite{baohan2014AAAI} explored a large set of
features and confirmed the effectiveness of mid-level representations
like SentiBank. In this work, we transfer emotion information learned
from the subset of the images in SentiBank \cite{Borth2013acmmm}
for the purpose of video emotion analysis.

The implicit approach for recognizing the emotional impact of a video
is to recognize emotions exhibited by viewers of that video. This
clever trick delegates the complex task of video understanding to
human viewers, thereby simplifying the problem. McDuff \emph{et al.}
\cite{RosalindPicard2015} analyzed facial expressions exhibited by
viewers of video advertisements recorded with webcams. Histogram of
Oriented Gradient \pl{(HOG)} features were extracted based on $22$
key points on the faces. Purchase intent is predicted based on the
entire emotion trajectory over time. Kapoor \etal \cite{Kapoor2007}
used video, skin conductance, and pressure sensors on the chair and
the mouse to predict frustration when a user interacted with an intelligent
tutoring system. However, the success of this approach depends on
the availability of a large number of human participants.

All previous work are limited as they aim to predict emotion or sentiment
classes present in the training set. In this work, we utilize knowledge
acquired from auxiliary images and text in order to identify emotion
classes unseen in the training set. In addition, we also investigate
related practical applications, previously unaddressed by the community,
like emotion-oriented video attribution and summarization.

\begin{figure*}
\begin{centering}
\includegraphics[width=0.9\textwidth]{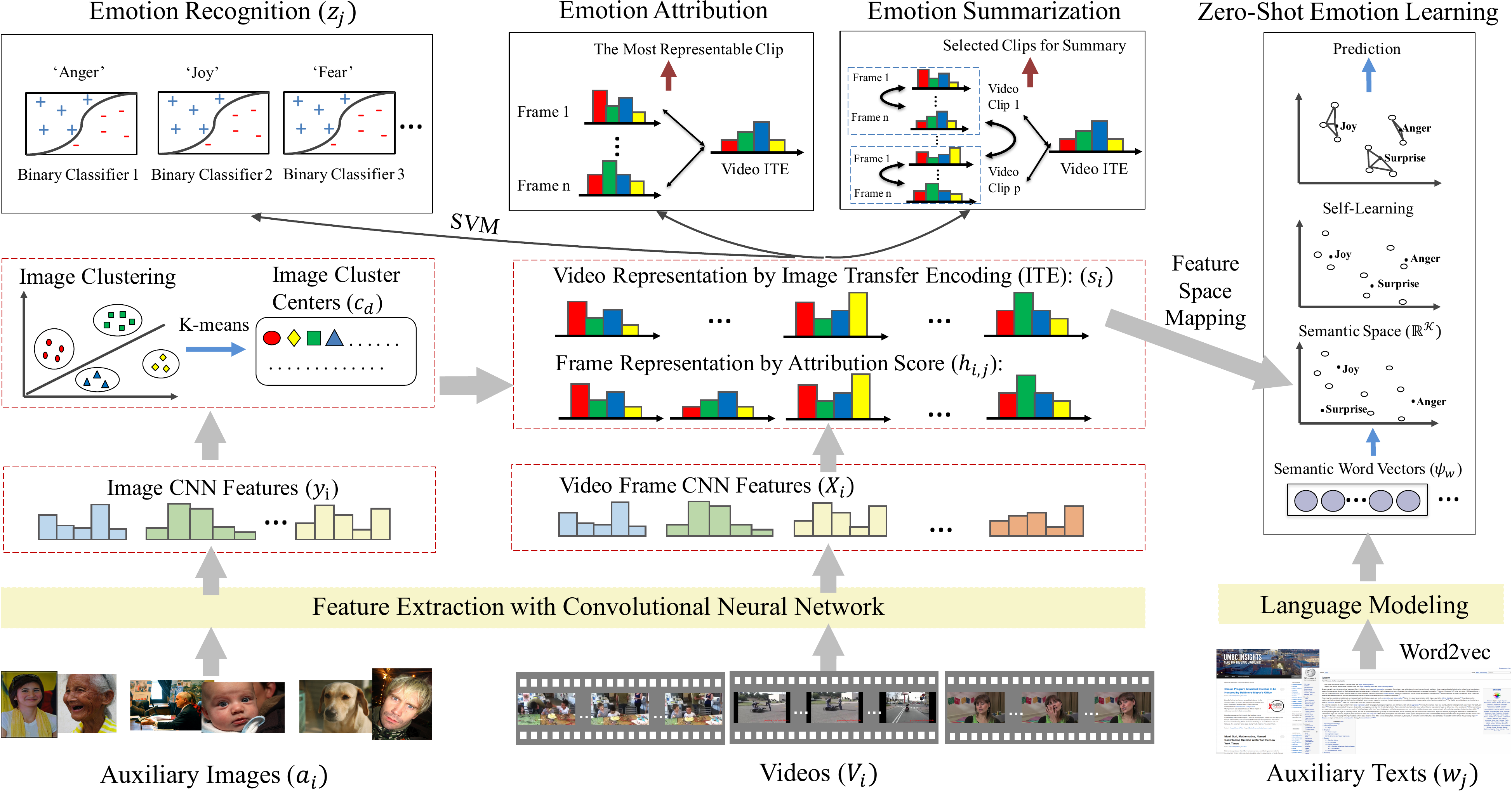} 
\par\end{centering}
\protect\protect\protect\caption{\label{fig:Overview-of-our} \textbf{An overview of our framework.}
Information from the auxiliary images (bottom left) is used to extract
an emotion-centric dictionary from CNN-encoded image elements, which
is subsequently used to encode video (bottom middle) and recognize
emotion (top left). The same encoding is used for emotion attribution
and summarization (top middle). Finally, information from a large
text corpora is utilized for zero-shot recognition of emotions, as
illustrated on the right.}
\end{figure*}

\subsection{Zero-shot Learning}

The tasks of identifying classes without any observed data is called
zero-shot learning~\cite{lampert13AwAPAMI}; its main challenge is
how to generalize the recognition models to identify novel object
categories without having access to any labelled instances. To solve
this problem, semantic attributes describing the properties across
object categories are used to transfer semantic knowledge from existing
to novel object classes~\cite{lampert13AwAPAMI,yanweiPAMIlatentattrib}.
However, these approaches require semantic attributes to be manually
defined and annotated; the availability of annotation limits their
scalability.

Recent work \cite{socher2013zero,frome2013devise} explore zero-shot
learning with representation of words as points in a multi-dimensional
vector space that is constructed from large-scale text corpora. The
intuition underlying this lexical representation is the distributional
hypothesis \cite{Harris1954}, which states that a word's meaning
is captured by other words that co-occur with it. This
representation has been demonstrated to exhibit generalization 
properties~\cite{distributedword2vec2013NIPS} and 
constructed vector space allows vector arithmetics
Our own experiments corroborate the benefits of such a model. For
example, when we add the vectors representing ``surprise'' and ``sadness'',
we obtain a vector whose the nearest neighbor under cosine similarity
is ``disappointment''. Adding vectors for ``joy'' and ``trust''
yields a vector whose nearest neighbor is ``love''.

In this paper, we explore zero-shot learning facilitated by a semantic
vector space that affords reasoning over emotion categories with vector
operations. The vector space and the operations capture knowledge
learned from text corpora. As we build regressors that project image
features into the text vector space, we are able to aggregate knowledge
learned from both images and text and identify semantic relations
between images and text. Subsequently, we can exploit the semantic
relations to recognize emotions unknown to the system. This knowledge
transfer framework is crucial to zero-shot emotion recognition. To
the best of our knowledge, this is the first work on zero-shot emotion
recognition.

\subsection{Video Emotion Attribution}

We define the novel task of \textit{emotion attribution} as attributing
the emotion of a video to its constituents such as frames or clips.
The video emotion attribution problem is inspired by sentiment attribution
in text \cite{Kotzias2014DeepMultinstance}. Besides the difference
in media (text \emph{vs.} video), our attribution problem also considers
multiple emotions whereas sentiment attribution considers only a binary
classification (positive \emph{vs.} negative).

\subsection{Video Summarization}

Video summarization has been studied for more than two decades. A
complete review is beyond the scope of this paper and we refer readers
to \cite{Truong:2007:VAS:1198302.1198305}. In broad strokes, we can
classify work on video summarization into two major categories: approaches
based on key frames\cite{hierarchical_video_summary,Alan_hanjalic_1998,DeMenthon98videosummarization,multi-view_yanwei_tr}
and approaches based on video skims~\cite{shotgraphSVM,event_driven_summary,DBLP:conf/mm/WangJCGDW14,fu2010summarize}.
Video summarization has been explored for various types of content,
including professional videos like movies or news reports \cite{shotgraphSVM,event_driven_summary,DBLP:conf/mm/WangJCGDW14},
egocentric videos \cite{story-driven-summary}, surveillance videos
\cite{fu2010summarize,multi-view_yanwei_tr}, and, to a lesser extent,
user-generated videos \cite{DBLP:conf/mm/WangJCGDW14,Gygli2014Creating}.

A diverse set of features have been proposed, including low-level
features such as visual saliency~\cite{Ma:2002:UAM:641007.641116}
and motion cues~\cite{shotgraphSVM,Pan2016Fast}, mid-level
information such as object trajectories~\cite{hierarchical_video_summary},
tag localization~\cite{event_driven_summary} and semantic recognition~\cite{DBLP:conf/mm/WangJCGDW14}.
Dhall and Roland \cite{ICPR12} considered smile/happy facial expressions.
User's spontaneous reactions such as eye movement, blink and face
expressions are measured by a Interest Meter system in \cite{peng2011editing}
and used for video summarization. However, none of these approaches
have considered video summarization based on more general video emotion
content. Such video emotion is an important cue for finding the most
``interesting'' or ``important'' video highlights. For example,
a good summary of a birthday party, or a graduation ceremony, should
capture emotional moments in the event. Not considering the valuable
emotion dimension in the video summarization task risks losing these
precious moments in the summary.

\subsection{Multi-Instance Learning }

The knowledge transfer approach adopted in this work is related to
multi-instance learning (MIL), which has been extensively studied
in the machine learning community. We hereby briefly review related
techniques. MIL refers to recognition problems where
each label is associated with a bag of instances, such as a bag of
video frames, rather than one data instance per label in the traditional
setting. 
Problem investigated in this work is intrinsically a multi-instance
learning case as each video consists of many frame instances with
possibly different emotions.

There are two main branches of MIL algorithms. The first branch
attempts to enable ``single-instance'' supervised learning algorithms
to be directly applicable to multi-instance feature bags. This branch
includes most of early works on MIL \cite{Multi-instance2013AI},
\cite{FG13} such as miSVM \cite{Andrews03supportvector}, MIBoosting
\cite{MIBoosting}, Citation-kNN \cite{Wang:2000:SMP:645529.757771},
MI-Kernel \cite{Gartner02multi-instancekernels}, among others \cite{Gu2016A,Gu2016Structural}.
These algorithms achieve satisfactory results in several applications
\cite{Ma2015Social,Wen2015A}, but most of them can
only handle small or moderate-sized data. In other words, they are
computationally expensive and cannot be applied to deal with large-scale
datasets.

The second branch of MIL adapt a multi-instance bag to a single data
instance in the original instance space. Popular algorithms include
constructive clustering based ensemble (CCE)~\cite{zhou2007}, multi-instance
learning based on the Fisher Vector representation (Mi-FV)~\cite{scalable_min_learning}
and multi-instance learning via embedded instance selection~\cite{MILES_multi-instance}.
Inspired by these works, we encode the video frame bags into single-instance
representations. It is worth noting our approach is different from
existing MIL algorithms because (1) we perform the encoding process
by using auxiliary image data, and demonstrate that transferring such
knowledge is important for video emotion analysis; (2) our emotion
recognition task is a multi-class multi-instance problem, while most
previous MIL algorithms aimed at binary classification.

\section{Approach}

In this section, we start by presenting the problem formulation and
common notations, and then discuss auxiliary image transfer encoding
and the three problems we tackle: zero-shot recognition, video emotion
attribution, and summarization. Figure \ref{fig:Overview-of-our}
shows an overview of our framework.

\subsection{Problem Setup}

We define our training video dataset with $n_{Tr}$ videos as: 
\[
Tr=\left\{ \left(V_{i},X_{i},\pmb s_{i},z_{i}\right)\right\} _{i=1,\cdots,n_{Tr}}
\]
where $V_{i}$ denotes the $i^{\text{th}}$ video, which is given
an emotion label $z_{i}$ and contains $n_{i}$ frames $f_{i,1},\ldots,f_{i,n_{i}}$.
Each frame is described by the feature vector $\mathbf{x}_{i,j}$.
As one video contains a set of features $X_{i}=\left\{ \mathbf{x}_{i,j}\right\} _{j=1,\cdots,n_{i}}$
and a single emotion label, this is a typical multi-instance learning
problem. Our MIL encoding process converts the bag of features $X_{i}$
into a video-level feature vector $\pmb s_{i}$ (see Section \ref{sub:ITE_subsec}).

In addition, we define a test set with $n_{Te}$ videos: 
\[
Te=\left\{ \left(V_{i},X_{i},\pmb s_{i},z_{i}^{\star}\right)\right\} _{i=1,\cdots,n_{Te}}
\]
where symbols are similarly defined except that $z_{i}^{\star}$ is
an emotion label in the test set. The notational difference is due
to the fact that in the zero-shot learning setting, no test labels
exist in the training set. Let $Z_{Tr}$ and $Z_{Te}$ denote emotion
labels in the training and test sets respectively, we have $Z_{Tr}\cap Z_{Te}=\varnothing$.

To enable knowledge transfer, we introduce a large-scale emotion-centric auxiliary
image set and a text dataset. We denote the auxiliary image sentiment
dataset as $A=\left\{ (a_{i},\pmb y_{i})\right\} _{i=1,\cdots,|A|}$
where $\pmb y_{i}$ is the feature vector of an image $a_{i}$. The
textual data are represented as a sequence of words $W=(w_{0},\ldots,w_{|W|})$,
$w_{j}\in\mathcal{V}$, where the vocabulary $\mathcal{V}$ is the
set of unique words. We learn a $\mathcal{K}$-dimensional embedding
$\pmb\psi_{w}$ for each word $w\in\mathcal{V}$, as detailed in Section
\ref{0-shot-emotion}.

In this paper, we extract image features $\mathbf{x}_{i,j}$ and $\pmb y_{i}$
with a deep Convolutional Neural Network (CNN) architecture, which
was recently shown to greatly outperform traditional hand-crafted
low-level features on several benchmark datasets, including MNIST
and ImageNet \cite{ImageNet2012}. Specifically, we retrain AlexNet
\cite{ImageNet2012} with all $2,600$ ImageNet classes and use the
activation of the seventh layer (``fc7'') as the feature vector
for each frame.

\subsection{Auxiliary Image Transfer Encoding (ITE) \label{sub:ITE_subsec}}

\subsubsection{The Encoding Scheme}

We utilize emotion information from a large-scale emotional image
dataset to encode each video into a video-level feature vector $\pmb s_{i}$ 
using a Bag-of-Words (BoW) representation. We learn a dictionary by
performing spherical k-means clustering \cite{hartigan1979algorithm}
on the auxiliary images, which finds $D$ spherical cluster centers
$\pmb c_{1}\ldots,\pmb c_{D}$. The similarity between a data point
$\pmb x_{i,j}$ and a cluster center $\pmb c_{d}$ is cosine similarity: 
\begin{equation}
\cos(\pmb x_{i,j},\pmb c_{d})=\frac{\pmb x_{i,j}^{\top}\pmb c_{d}}{\|\pmb x_{i,j}\|\|\pmb c_{d}\|}.\label{eq:similarity}
\end{equation}
We use $D$ cluster centers from a dictionary to encode a video into
a $D$-dimensional BoW feature vector. Recall that a video $V_{i}$
contains $n_{i}$ frames and corresponding features $X_{i}=\left\{ \mathbf{x}_{i,j}\right\} _{j=1,\cdots,n_{i}}$.
For each frame, we identify its $K$ nearest cluster centers. We can
compute the assignment variables $\gamma_{i,j,d}$ as if we assign
frame $f_{i,j}$ to the $d^{\text{th}}$ cluster: 
\begin{equation}
\gamma_{i,j,d}=\begin{cases}
\begin{array}{cc}
1 & if\,\pmb c_{d}\in K\text{-NN}\left(\mathbf{x}_{i,j}\right),\\
0 & otherwise,
\end{array}\end{cases}\label{eq:kmeans-define}
\end{equation}
where $K\text{-NN}\left(\mathbf{x}_{i,j}\right)$ denotes the spherical
$K$ nearest neighbours to $\mathbf{x}_{i,j}$ from all cluster centers.
The video-level encoding $\pmb s_{i}$ is the accumulation 
of the frames; the $d^{\text{th}}$ dimension of $\pmb s_{i}$ is
computed as: 
\begin{equation}
s_{i,d}=\sum_{j=1}^{n_{i}}\gamma_{i,j,d}\cdot\cos\left(\mathbf{x}_{i,j},\pmb c_{d}\right).\label{eq:video-level-sim}
\end{equation}

\subsubsection{Rationale for ITE}


We utilize emotional information from
a large-scale emotional image dataset to help encode the video content,
i.e. ITE. This can be intuitively explained from the perspective of
entropy. A dictionary built from the auxiliary emotion-related images
can efficiently encode a video frame with emotion information as a
sparse vector which concentrates on a few dimensions. In comparison,
a frame without emotion information will likely be encoded less efficiently,
producing a denser vector with small values in many dimensions. As
a result, a non-emotional frame will have higher entropy than the
emotional frame, and hence less impact on the resulted BoW representation.

Our encoding scheme ITE also differs from the standard BoW \cite{sivic2003video_google}
and soft-weighting BoW \cite{tmm10:yjiang,Zhou2016Effective}. The standard
BoW encodes local descriptors, such as SIFT and STIP, which requires
a dictionary orders of magnitude greater than our frame set.\textcolor{blue}{{}
}Thus directly using standard BoW \cite{sivic2003video_google} on
our problem will make the generated video-level features too sparse
to be discriminative. Second, soft-weighting encoding, as a sophisticated
and refined version of standard BoW, weight the significance of visual
words with decaying weights on more cluster center bins.
In contrast, due to the diverse nature of emotions (as explained in
Section \ref{section:psyc-theories}), one single video frame may
equally evoke multiple emotions from viewers. We allow one feature
vector $\mathbf{x}_{i,j}$ to equally contribute to multiple encoding
bins (Eq. \ref{eq:video-level-sim}).

\subsection{Zero-Shot Emotion Recognition}

\label{0-shot-emotion} Canonical emotion theories \cite{nature_emotion}
often define a fixed number of prototypical emotions. However, recent
research \pl{\cite{Barrett2006,Lindquist2012}} highlights differences
within each emotion category and argues that emotions are more diverse
than previously imagined. This raises an interesting question: can
we identify emotions that are not in our training set purely from
their class labels? This is the zero-shot recognition problem.

To address this difficult challenge, we relate emotion class labels
we have not seen before to the class labels we have seen. We learn
a distributed representation for class label words from an auxiliary
corpora of text containing emotional data, utilizing the linguistic
intuition that words appear in similar contexts usually have similar
meaning~\cite{Harris1954}. The distributed representations are embedded
in a low-dimensional space $\mathbb{R}^{\mathcal{K}}$ in which emotion
class labels can be related to each other.

Following Mikilov \textit{et al.} \cite{distributedword2vec2013NIPS},
we learn the distributed representation by predicting from each word
its context words. Given a word $w_{t}$ and its surrounding context
words $(w_{t-M},\cdots,w_{t-1},w_{t+1},\cdots,w_{t+M})$ within a
window of size $2M$, we maximize the log likelihood of context words
conditioned on $w_{t}$: 
\begin{equation}
\max\sum_{t}\sum_{-M\leq j\leq M,j\neq0}\log p(w_{t+j}|w_{t})\label{eq:learningwordspace-1}
\end{equation}
We represent every unique word $w$ in the vocabulary $\mathcal{V}$
as a vector $\pmb\psi_{w}\in\mathbb{R}^{\mathcal{K}}$ and parameterize
the above likelihood: 
\begin{equation}
p(w_{t+j}|w_{t})\propto\exp\left({\pmb\psi}_{w_{t+j}}^{\top}{\pmb\psi}_{w_{t}}\right).\label{eq:learningwordspace-2}
\end{equation}
Directly optimizing Eq. \ref{eq:learningwordspace-1} is intractable
because computing the probability in Eq. \ref{eq:learningwordspace-2}
requires a summation over all words in the vocabulary. As an approximation,
we use the negative sampling technique, which samples a few negative
examples $w_{1}^{\prime},\cdots,w_{m}^{\prime}$ that do not appear
in the context window, and maximizes: 
\begin{equation}
\sum_{-M\leq j\leq M,j\neq0}\log\sigma\left({\pmb\psi}_{w_{t+j}}^{\top}{\pmb\psi}_{w_{t}}\right)-\sum_{1\le j\le m}\log\sigma\left({\pmb\psi}_{w_{j}^{\prime}}^{\top}{\pmb\psi}_{w_{t}}\right)
\end{equation}

Training of the above model yields embeddings for each word in the
vocabulary. We can then train a regressor $g(\cdot)$ from video-level
features $\pmb s_{i}$ to the embedding of its class label word (e.g.,
joy, sadness). In this work, we train a support vector regressor with
a linear kernel for each dimension of the word vector.

However, regressors trained on the training set may generalize poorly
to test classes that do not exist in the training set. This is mainly
because the distribution of visual features in different classes differ.
For example, videos of joy usually have positive frames with bright
light and smiling faces, while a sad video would typically contain
dark colors and people crying. Thus, the relation between video features
and the class label's embedding may vary for different classes.

To alleviate this generalization problem, we take inspiration from
the Rocchio algorithm in information retrieval~\cite{vector_space_classification,yanweiPAMIlatentattrib},
and use more relevant testing instances to update the query prototypes
for better classification accuracy. We thus apply Transductive 1-Step
Self-Training (T1S) to adjust the word vector of unseen emotion classes.
Let $Z_{Tr}$ and $Z_{Te}$ denote emotion label words in the training
and test sets respectively. For a test class $z_{i}^{\star}\in Z_{Te}$
that is previously unseen and its distributed representation ${\pmb\psi}_{z_{i}^{\star}}$,
we relate it to $k$ nearest video neighbors in the test set. We compute
a smoothed version $\bar{{\pmb\psi}}_{z_{i}^{\star}}$: 
\begin{equation}
\bar{{\pmb\psi}}_{z_{i}^{\star}}=\frac{1}{K}\sum_{g\left(\pmb s_{k}\right)\in K\text{-NN}\left({\pmb\psi}_{z_{i}^{\star}}\right),V_{k}\in Te}^{}g\left(\pmb s_{k}\right),\label{eq:self-training}
\end{equation}
where $K\text{-NN}\left(\cdot\right)$ denotes the set of spherical
$K$ nearest neighbors in the semantic space. Eq. (\ref{eq:self-training})
aims to transductively ameliorate visual differences by averaging
$\pmb\psi_{z_{i}^{\star}}$ with nearest test instances. Here, to
prevent semantic drift caused by self-training, we only perform self-training
once.

After obtaining $\bar{\pmb\psi}_{z_{i}^{\star}}$ for all unseen classes
$z_{i}^{\star}\in Z_{Te}$, we can do nearest-neighbor classification
in the vector space. Given a test video $V_{j}$ and its video level
features $\pmb s_{j}$, its class label ${z}_{j}^{\star}$ can be
estimated as: 
\begin{equation}
\hat{z}_{j}^{\star}=\argmax_{z^{\star}\in Z_{Te}}\,\cos\left(g\left(\pmb s_{j}\right),\bar{\pmb\psi}_{z^{\star}}\right).\label{eq:self-training-1-step}
\end{equation}

Compared with the zero-shot learning algorithm in \cite{yanweiPAMIlatentattrib},
we skip the intermediate level of latent attributes and directly apply
the 1-step self-training in the semantic word vector space. In addition,
we use cosine similarity as the metric rather than the Euclidean distance
since the semantic word vectors are intrinsically directional and
cosine similarity is a better metric used in \cite{distributedword2vec2013NIPS,yanwei_ZSL_pami}.
The process is summarized in Algorithm \ref{Alg: T1S}.

\begin{algorithm}[t] 
\small 	
\caption{ Pseudo-code describing the T1S algorithm.} 	\label{Alg: T1S}
 	 \begin{algorithmic}[1]  	 
	\REQUIRE ~~\\ 	 	 
$\bullet$ $C:$ an auxiliary text dataset . \\ 	
	$\bullet$ $Tr:$ training video set;\\ 	 	
$\bullet$ $Te:$ testing video set. \\ 	
$\bullet$ $Z_{Tr}\cap Z_{Te}=\varnothing$.  \\
	\STATE Train the word2vec language model with the large-scale auxiliary text dataset  $\leftarrow$  Eq (\ref{eq:learningwordspace-1}) 	\STATE Project emotion words of training and test sets to embeddings $\pmb \psi_{w_{Tr}}$ and $\pmb \psi_{w_{Te}}$; 	 
	\STATE Train regressors from video features to class label's embeddings
	 \STATE Perform zero-shot emotion recognition $\leftarrow$ Eq (\ref{eq:self-training}) and Eq (\ref{eq:self-training-1-step}). 	 	 	 \end{algorithmic} 
\end{algorithm}

\subsection{Video Emotion Attribution}

Emotion attribution aims to identify the contribution of each frame
to the video's overall emotion. Emotion attribution can help find
\textit{video highlights} \cite{Alan_hanjalic_1998}, which are defined
as interesting or important events in the video. Generally, the concepts
of ``interesting'' and ''important'' may be variable for different
video domains and applications, such as the scoring of a goal in soccer
videos, applause and cheering in talk-show videos, and exciting speech
in presentation videos. Nevertheless, most of these ``interesting''
or ``important'' video events convey very strong video emotions,
thus providing important signal for highlighting the core parts of
the whole video.

Formally, for a video $V_{i}$ containing a sequence of frames $f_{i,1},\ldots,f_{i,n_{i}}$,
we want to find the frames that substantially contribute to the overall
video emotion. Using the ITE technique described in Section \ref{sub:ITE_subsec},
we can encode a frame $f_{i,j}$ as its similarity to $D$ cluster
centers: 
\begin{equation}
\pmb h_{i,j}=\left[\cdots,\mathbf{\gamma}_{i,j,d}\cdot\cos(\mathbf{x}_{i,j},\pmb{c}_{d}),\cdots\right]_{1\le d\le D}\label{eq:emotional_score_vector}
\end{equation}
where $\mathbf{\gamma}_{i,j,d}$ is defined in Eq (\ref{eq:kmeans-define}).
The vector $\pmb h_{i,j}$ uses the auxiliary image dataset $A$ to
evaluate the emotions in the $j^{\text{th}}$ frame.
Note that in this case $\sum_j \pmb h_{i,j} = \pmb s_i$.

The video emotion attribution can then be formulated as measuring
the similarity between the video-level emotion vector and the frame-level
vectors. Specifically, the attribution score of the $j^{\text{th}}$
video frame is computed as the cosine similarity between the video-level
feature $\pmb s_{i}$, and frame-level feature $\pmb h_{i,j}$. We
thus find the frame that contributes the most to the overall emotion
of $V_{i}$ by 
\begin{equation}
\argmax_{j\in[1,...,n_{i}]}\,\cos\left(\pmb s_{i},\pmb h_{i,j}\right).\label{eq:maximum_attribution}
\end{equation}

The emotion attribution procedure can also be extended to a list of
pre-partitioned video clips $\left\{ E_{1},\ldots E_{P}\right\} $
by using clip-level feature $\pmb h_{p}$: 
\begin{equation}
\pmb h_{p}=\left[\cdots,\sum_{f_{j}\in E_{p}}\mathbf{\nu}_{i,j,d}\cdot\cos(\mathbf{x}_{i,j},\pmb{c}_{d}),\cdots\right]_{1\le d\le D}\label{eq:emotional_score_vector_clip_level}
\end{equation}

\subsection{Emotion-Oriented Video Summarization}

One important problem that is enabled by emotion attribution is video
summarization. Leveraging our proposed technique, here we present
a video summarization method for preserving the emotional content
in a video and balancing it against information coverage.

We summarize a video by extracting a number of key frames from it.
Let $U_{i}$ denote this set of key frames for video $V_{i}$, we
select  highest scored frames according to the following: 

\begin{equation}
U_{i}=\argmax_{f_{j}\in V_{i}}\cos\left(\pmb s_{i},\pmb h_{i,j}\right)+\lambda\sum_{f_{k}\in V_{i}}\cos\left(\mathbf{x}_{i,j},\mathbf{x}_{i,k}\right),\label{eq:key-frame_summary}
\end{equation}
where $\lambda$ is a weight parameter, and the second term $\cos\left(\mathbf{x}_{i,j},\mathbf{x}_{i,k}\right)$
rewards key frames that are the most similar to other frames in the
same video, which means that the selected frames are representative
of the entire video. Note that (1) the cosine similarity 
in the first term is computed using ITE, while in the second term the
similarity is defined directly in the feature space; 
(2) We empirically set $\lambda=1$ to equally
consider both emotion content and representativeness of the video. 

Comparing with previous work \cite{Truong:2007:VAS:1198302.1198305,Alan_hanjalic_1998,multi-view_yanwei_tr,fu2010summarize,DBLP:conf/mm/WangJCGDW14},
Eq (\ref{eq:key-frame_summary}) considers the summary of both video
highlights (by the first term for emotion attribution) and information
coverage (by the second term for eliminating redundancy and selecting
information-centric frames/clips). Thus our method can produce a condensed,
succinct and emotion-rich summary which can facilitate the browsing,
retrieval and storage of the original video content. Particularly,
our summary results are more emotionally interpretable due to the
emotion attribution.

\section{Experiments}

\subsection{Datasets and Settings}

\label{sec:implementation_details} We utilize three video emotion
datasets for evaluation. Among them, the VideoStory-P14 and YF-E6
datasets are introduced by us and will be made available to the community.

\vspace{0.1in}

\begin{table}
\begin{centering}
\textcolor{blue}{}%
\begin{tabular}{|c|c|c|c|}
\hline 
 & $P_{0}$  & $P_{e}$  & Kappa \tabularnewline
\hline 
\hline 
YouTube-24  & $0.74$  & $0.31$  & $0.62$ \tabularnewline
\hline 
YF-E6 & $0.82$  & $0.33$  & $0.73$ \tabularnewline
\hline 
\end{tabular}
\par\end{centering}
\caption{\label{tab:kappa}Cohen's kappa score of the annotations
of two dataset annotated in this paper. $P_{0}$ is the relative observed
agreement among annotators. And $P_{e}$ is the hypothetical probability
of chance agreement. For YouTube-24 the annotators were tasked with 
sub-annotating each of the videos in $8$ emotional categories into $3$ 
additional sub-categories; this is the reason $P_{e}$ is relatively high, as
chance agreement is only among the $3$ sub-categories.}

\end{table}

\noindent \textbf{YouTube emotion datasets \cite{baohan2014AAAI}.}
The YouTube dataset contains 1,101 videos annotated with $8$ basic
emotions from the Plutchik's Wheel of Emotions. To facilitate the
zero-shot emotion recognition task, we re-annotate the videos into
$24$ emotions, by adding $3$ variations to each basic emotion according
to Plutchik's definition. For example, we split the \emph{joy} class
into \emph{ecstasy}, \emph{joy} and \emph{serenity} according to arousal.
We use the short-hand \textbf{YouTube-8} and \textbf{YouTube-24} for
the original and re-annotated datasets respectively. To
annotate the \textbf{YouTube-24}  dataset,
each video was labeled by 5 annotators using majority vote with a
high Cohen's kappa score\footnote{Cohen's kappa coefficient is a metric for measuring
the inter-rater agreement for qualitative items (i.e. annotations).} \cite{Smeeton1984Early} of $0.62$ (in Table \ref{tab:kappa}). 

We hereby report statistics for emotions included in the datasets.
YouTube-8 dataset contains 101 videos labeled as \emph{anger}, 101
as \emph{anticipation}, 115 \emph{disgust}, 167 \emph{fear}, 180 \emph{joy},
101 \emph{sadness}, 236 \emph{surprise}{} and 100 \emph{trust}. YouTube-24
has 36 videos labeled as \emph{anger}, 33 \emph{annoyance}, 32 \emph{rage},
44 \emph{anticipation}, 32 \emph{interest}, 25 \emph{vigilance}, 42
\emph{boredom}, 64 \emph{disgust}, 9 \emph{loathing}, 12 \emph{apprehension},
79 \emph{fear}, 76 \emph{terror}, 23 \emph{ecstacy}, 76 \emph{joy},
81 \emph{serenity}, 27 \emph{grief}, 11 \emph{pensiveness}, 63 \emph{sadness},
29 \emph{amazement}, 59 \emph{distraction}, 148 \emph{surprise}, 39
\emph{acceptance}, 26 \emph{admiration}, 35 \emph{trust}. 

\vspace{0.1in}
\noindent \textbf{YouTube/Flickr-EkmanSix (YF-E6) dataset.} As discussed
in the related work section, Ekman \cite{nature_emotion} found a
high agreement of emotions across cultures and proposed $6$ basic
emotion types. We collect the YF-E6 emotion dataset using the $6$
basic emotion type as keywords on social video-sharing websites including
YouTube and Flickr, leading to a total of $3000$ videos. The dataset
is labeled through crowdsourcing by 10 different annotators (5 males
and 5 females), whose age ranged from $22$ to $45$. Annotators were
given detailed definition for each emotion before performing the task.
Every video is manually labeled by all the annotators. A video is
excluded from the final dataset when over half of annotations are
inconsistent with the initial search keyword. Due to high agreement
of Ekman emotions, we observe very high consistency of the annotations:
$85\%$ videos were given the same label by 7 or more annotators with
the high kappa score $0.73$ (in Table \ref{tab:kappa}). The final
dataset comprises $1,637$ videos across the $6$ emotion classes,
with an average duration of 112 seconds. Specifically, the YF-E6 dataset
contains 225 \emph{anger}, 239 \emph{disgust}, 287 \emph{joy}, 221
\emph{sadness}, and 360 \emph{surprise} videos.

 \vspace{0.1in}
\noindent \textbf{The VideoStory-P14 dataset}. The VideoStory-P14
dataset is derived from the recently proposed VideoStory dataset \cite{FewShot2014ACMMM}.
We use all the keywords of the Plutchik's Wheel of Emotions \cite{Plutchik1980}
to query the VideoStory dataset in terms of its video captions. Emotion
keywords are matched against all the words in the video's caption.
This leads to a set of $626$ videos belonging to $14$ emotion classes.
The dataset contains 83 videos labeled as \emph{anger},
30 as \emph{annoy}, 27 \emph{aggressive},
119 \emph{rage}, 28 \emph{interest},
14 \emph{disgust}, 29 \emph{distract},
16 \emph{fear}, 23 \emph{terror},
67 \emph{love}, 80 \emph{joy},
81 \emph{surprise}, 11 \emph{submission},
18 \emph{trust}. 

 \vspace{0.1in}
\noindent \textbf{Auxiliary emotional image and text datasets.} From
the Flickr image dataset \cite{Borth2013acmmm}, we select as the
auxiliary image data a subset of $110K$ images of Adjective-Noun
Pairs (ANPs) that have top ranks with respect to the emotions (see
Table 2 in \cite{Borth2013acmmm}). These images are clustered into
$2,000$ clusters (i.e. $D=2000$ in Eq (\ref{eq:similarity})). As
shown in \cite{distributedword2vec2013NIPS}, the large-scale text
data can greatly benefit the trained language model. We \pl{train
the Skip-gram model (Eq \ref{eq:learningwordspace-2})} on a large-scale
text corpora, which includes around 7 billion words from the UMBC
WebBase (3 billion words), the latest Wikipedia articles (3 billion
words) and some other documents (1 billion words). The trained model
contains roughly 4 million unique words, bi-gram and tri-gram phrases
(i.e., $|\mathcal{V}|\thickapprox4\,$ million). Most of the documents
are formal texts which have clear definitions, descriptions and usage
of the emotion and sentiment related words.

 \vspace{0.1in}
\noindent \textbf{Experimental settings.} Each video is uniformly
sampled at $5$ frame increments for feature extraction to reduce
the computational cost. The dimension of the real-valued semantic
vectors $\mathbf{\pmb\psi}_{w}$ (Eq (\ref{eq:learningwordspace-2}))
is set to $500$ to balance computational cost of training $\mathbf{\pmb\psi}_{w}$
from large-scale text corpora and the effectiveness of the syntactic
and semantic regularities of representations \cite{distributedword2vec2013NIPS}.
Our AlexNet CNN model is trained by ourselves using $2,600$ ImageNet
classes with the Caffe toolkit \cite{jia2014caffe}, and we use the
$4,096$-dimensional activations of the 7th fully-connected layer
after the Rectified Linear Units (\emph{i.e.} $fc7$) as features.
The number of nearest neighbors in Eq (\ref{eq:kmeans-define}) is
empirically set to $10\%$ of the image clusters (i.e. $K=D/10$),
which balances the computational cost with a good representation in
Eq (\ref{eq:video-level-sim}).

\subsection{Supervised Emotion Recognition}

\noindent To illustrate benefits of our ITE encoding scheme, we first
perform supervised emotion recognition with a support vector machines
(SVM) classifier with chi-square kernel. We compare our method with
the following alternative baselines.

\vspace{0.1in}
\noindent \textbf{MaxP~\cite{DBLP:journals/jmlr/LiuWZ12}.} The
instance-level classifiers are trained using the labels inherited
from their corresponding bags. These classifiers can be used to predict
instance labels of testing videos. The final bag labels are produced
by majority vote of instance labels. This method is a variant of the
Key Instance Detection (KID) \cite{DBLP:journals/jmlr/LiuWZ12} in
multi-class multi-instance setting.

 \vspace{0.1in}
\noindent \textbf{AvgP~\cite{DBLP:journals/corr/XuYH14}.} We average
the frame-level image features of one video as video-level feature
descriptions for classification. For the $i^{\text{th}}$ video, its
average pooling feature is computed as $\frac{1}{n_{i}}\sum_{j=1}^{n_{i}}\mathbf{x}_{i,j}$.
The average pooling is the standard approach of aggregating frame-level
features into video-level descriptions as mentioned in \cite{DBLP:journals/corr/XuYH14}.

 \vspace{0.1in}
\noindent \textbf{Mi-FV~\cite{scalable_min_learning}.} MIL bags
of training videos are mapped into a new bag-level Fisher Vector representation.
Mi-FV is able handle large-scale MIL data efficiently.

 \vspace{0.1in}
\noindent \textbf{CCE~\cite{zhou2007}.} The instances of all training
bags are clustered into $b$ groups, and each bag is re-represented
by $b$ binary features, where the value of the
$i^{\text{th}}$ feature is $1$ if the concerned bag has instances
falling into the $i^{\text{th}}$ group and $0$ otherwise. This is
essentially a simplified version of our ITE method encoded by training
instances only.

\vspace{0.1in}
\noindent The linear kernel is used for Mi-FV and MaxP due to the
large number of samples/dimensions, and the Chi-square kernel\footnote{The RBF kernel is also evaluated but shows slightly lower performance
than that of the Chi-square kernel.} is used for others. A binary two-class SVM model is trained for each
emotion class separately. The key parameters are selected by 3-fold
cross-validation.

\begin{figure*}[t]
\centering \includegraphics[scale=0.32]{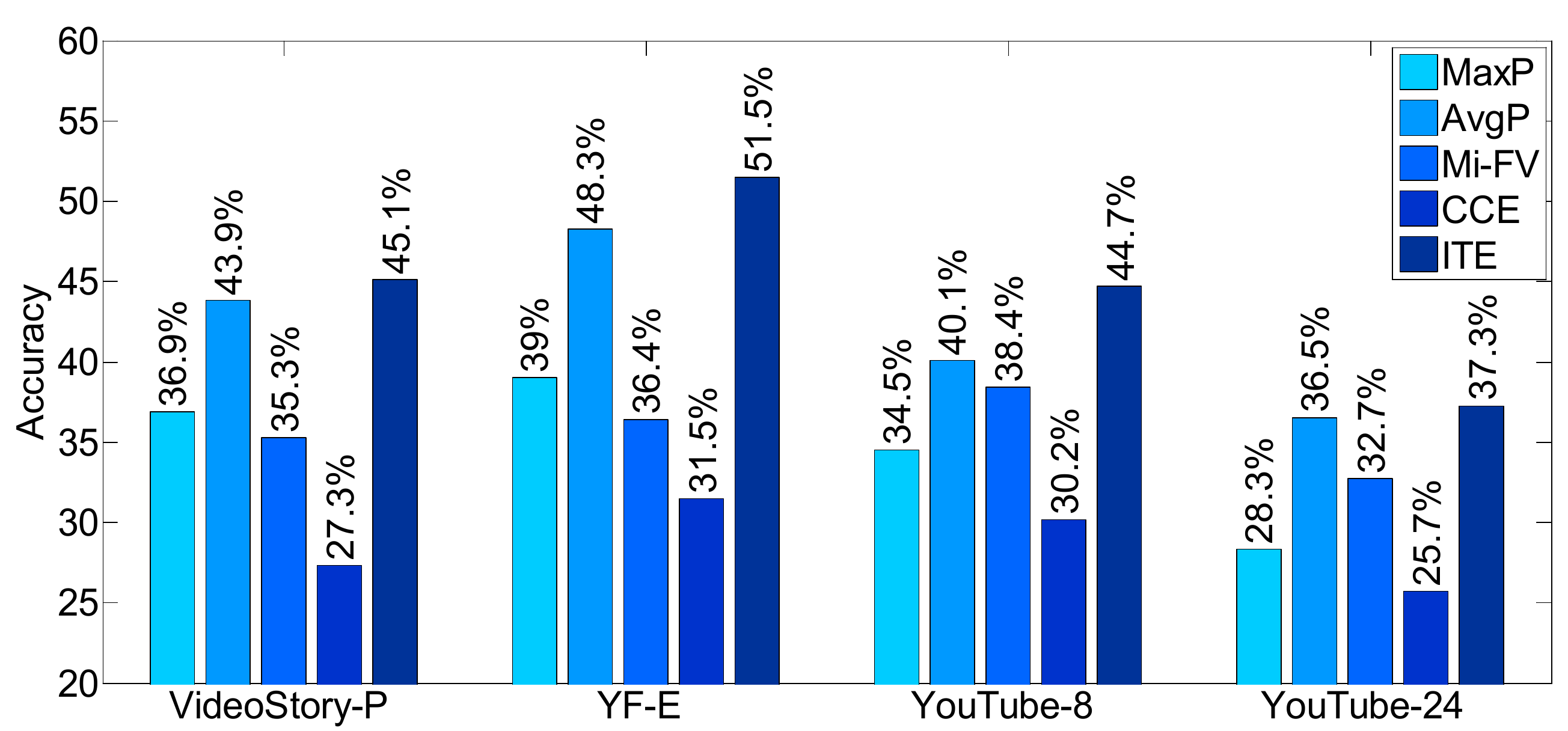}
\protect\protect\caption{\label{tab:Supervised-Learning-on} Average accuracy of supervised
emotion recognition.}
\end{figure*}

\begin{figure}[t]
\centering \includegraphics[scale=0.5]{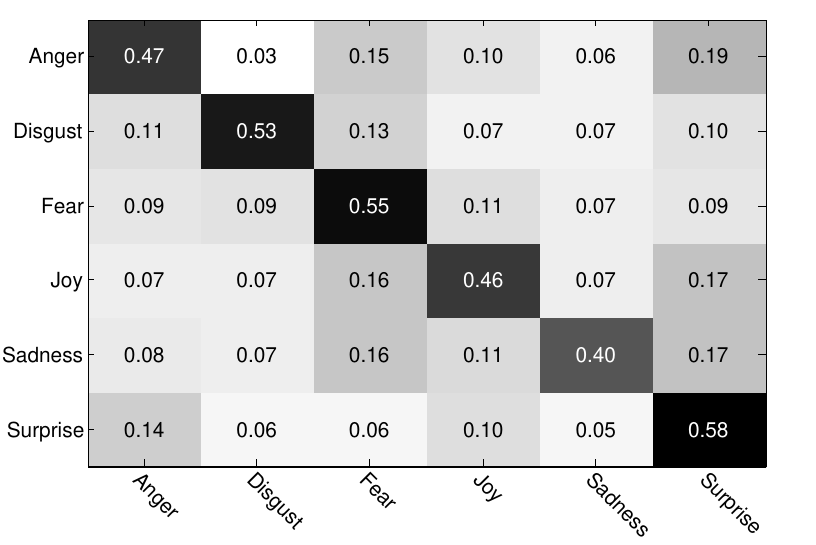} \includegraphics[scale=0.5]{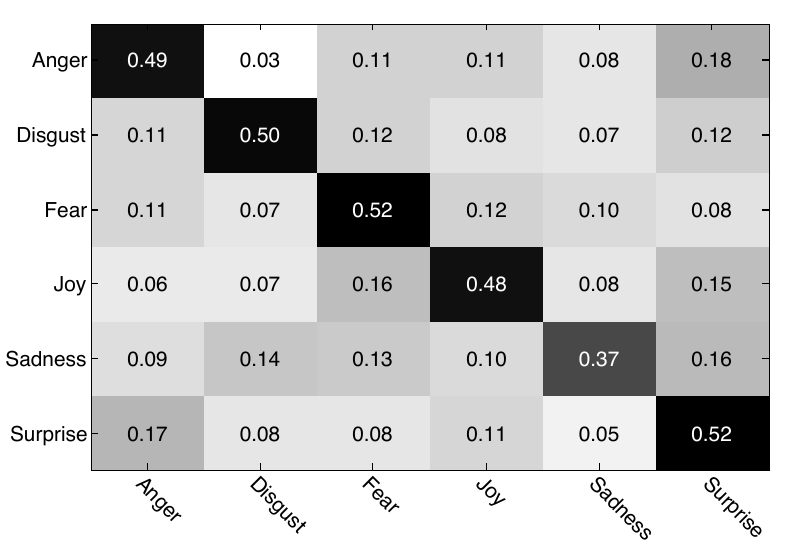}
\caption{\label{fig:confusion-matrix-supervised}Confusion matrices for supervised
learning on the YF-E6 dataset using our ITE encoding (top) and the
AvgP method (bottom).}
\end{figure}

\begin{figure*}[t]
\begin{centering}
\includegraphics[scale=0.32]{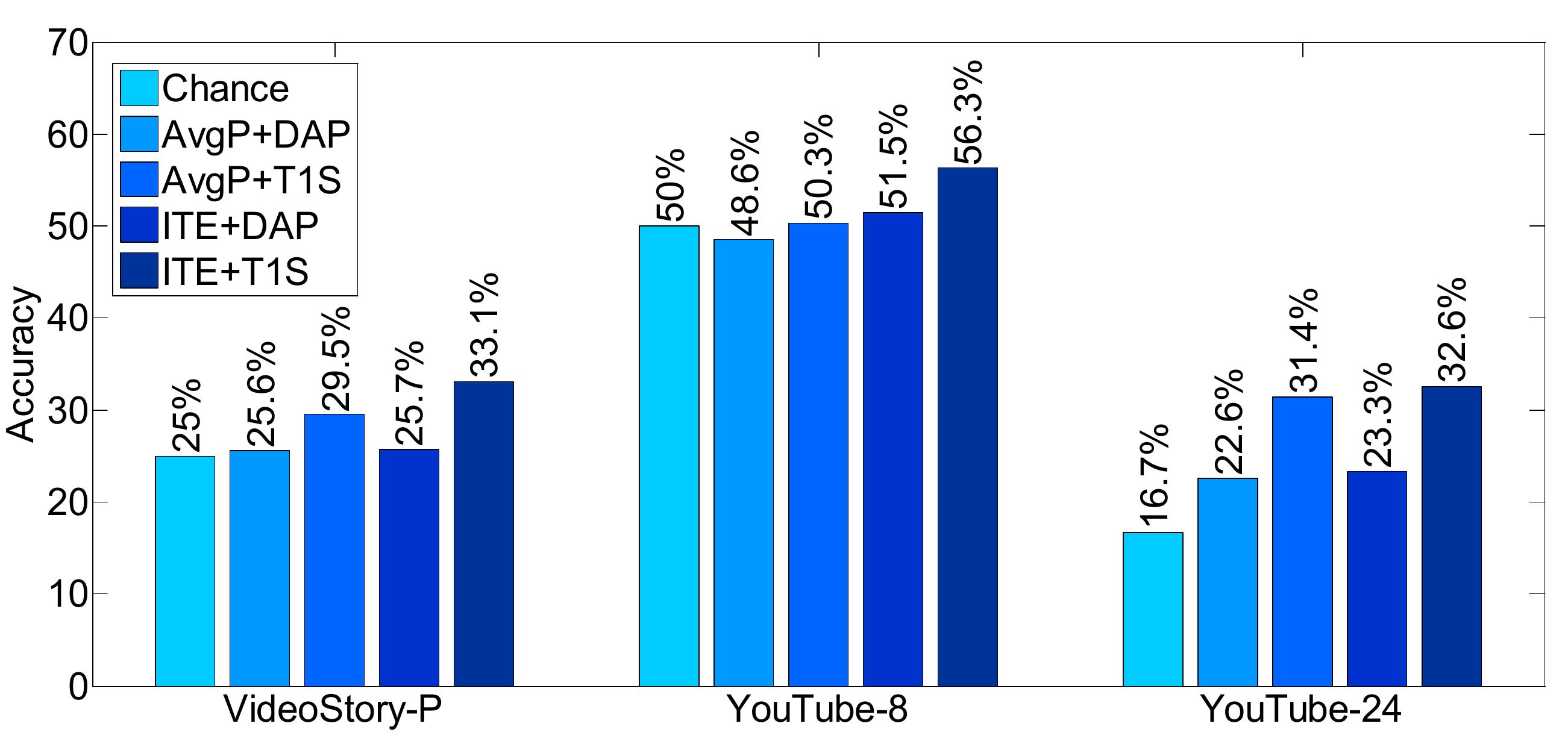} 
\par\end{centering}
\protect\protect\caption{\label{tab:Zero-shot-Learning-results}Average accuracy
of zero-shot emotion recognition on YouTube and VideoStory-P14 datasets.}
\end{figure*}

The experimental results are shown in Figure \ref{tab:Supervised-Learning-on},
which clearly demonstrate that our ITE method significantly outperforms
the four alternatives on all three datasets. This validates the effectiveness
of our method in generating a better video-level feature representation
based on the auxiliary images. In particular, the improvement of ITE
over CCE and Mi-FV verifies that the knowledge transferred from the
auxiliary emotional image dataset is probably more critical than that
existing in the training video frames. This supports our argument
that most of the frames of these videos have no direct relation to
the emotions expressed by the videos, and underscores the importance
of knowledge transfer.{} Particularly, we use the same training/testing
split as \cite{baohan2014AAAI} on \textbf{YouTube-8}{} dataset.
The AvgP result $40.1\%$ is comparable with the result $41.9\%$
in \cite{baohan2014AAAI} with the method of combining different types
of hand-crafted visual features with the state-of-the-art multi-kernel
strategy. This validates that the performance of the deep features
we use can match that of multi-kernel combination of hand-crafted
features.

Our ITE results show $1.2$, $3.2$, $4.6$, $0.8$ absolute percentage
points improvement over AvgP on VideoStory-P14, YF-E6, Youtube-8 and
YouTube-24 video dataset. This further validates the effectiveness
of our method. In particular, we found that (1) a portion of videos
in VideoStory-P14 dataset are surveillance videos (from VideoStory
dataset), which have very poor visual quality and thus make all the
methods fail. This explains the slightly lower improvement margin
of ITE over AvgP on VideoStory-P14 dataset. (2) Also note that Youtube-24
is a re-annotated version of Youtube-8 and thus it is even harder
because in contains larger number of classes and fewer training instances
per class. While it is difficult for all the methods to classify emotions,
our ITE results are still the best among the competitors.

One should notice that CCE has the worst performance. CCE re-encodes
the multi-instances into \emph{binary} representations by ensemble
clustering. Such representations may have better performance than
the hand-crafted features used in \cite{zhou2007}, but they cannot
beat the recently proposed deep features, which have been shown to
be able to extract higher level information \cite{ImageNet2012}.
In other words, the re-encoding process of CCE loses discriminative
information gained from the deep features, and is therefore unsuited
for the task.

In addition, Mi-FV and MaxP have similar performance: MaxP is slightly
better on VideoStory-P14, YF-E6 and Mi-FV is slightly better on YouTube.
However, the results of Mi-FV and MaxP are much worse than those of
AvgP. These differences can be explained by the different choices
of kernels. We validate that the AvgP with linear SVM classifier has
similar performance (with a variance of $2\%$) as MaxP and Mi-FV.
Nevertheless, due to high dimensions of Fisher Vectors and large amount
of training instances in MaxP, nonlinear kernels will introduce prohibitive
computational cost. Thus, in subsequent experiments, we use AvgP as
the main alternative baseline to ITE, since other alternatives do
not demonstrate competitive advantages. We illustrate the confusion
matrix in Fig.~\ref{fig:confusion-matrix-supervised}. The matrices
of ITE shows clear diagonal structure and the results are better than
AvgP in most of classes.

Some qualitative results of supervised emotion prediction are shown
in Figure \ref{fig:superised-example}. In the successful cases, test
videos share visual characteristics with auxiliary image dataset,
such as \emph{bright light} and {\emph{smiling faces} in the ``joy''
category. The ``anger'' videos are wrongly classified as ``fear''.
Comparing with ``anger'', the ``fear'' category is more highly
correlated with \emph{dark lightning} and \emph{screaming faces} which
are visually dominant in the failure case. The video wrongly labeled
as ``joy'' has festive colors which resemble a Christmas tree.

\begin{figure}[t]
\begin{centering}
\includegraphics[scale=0.35]{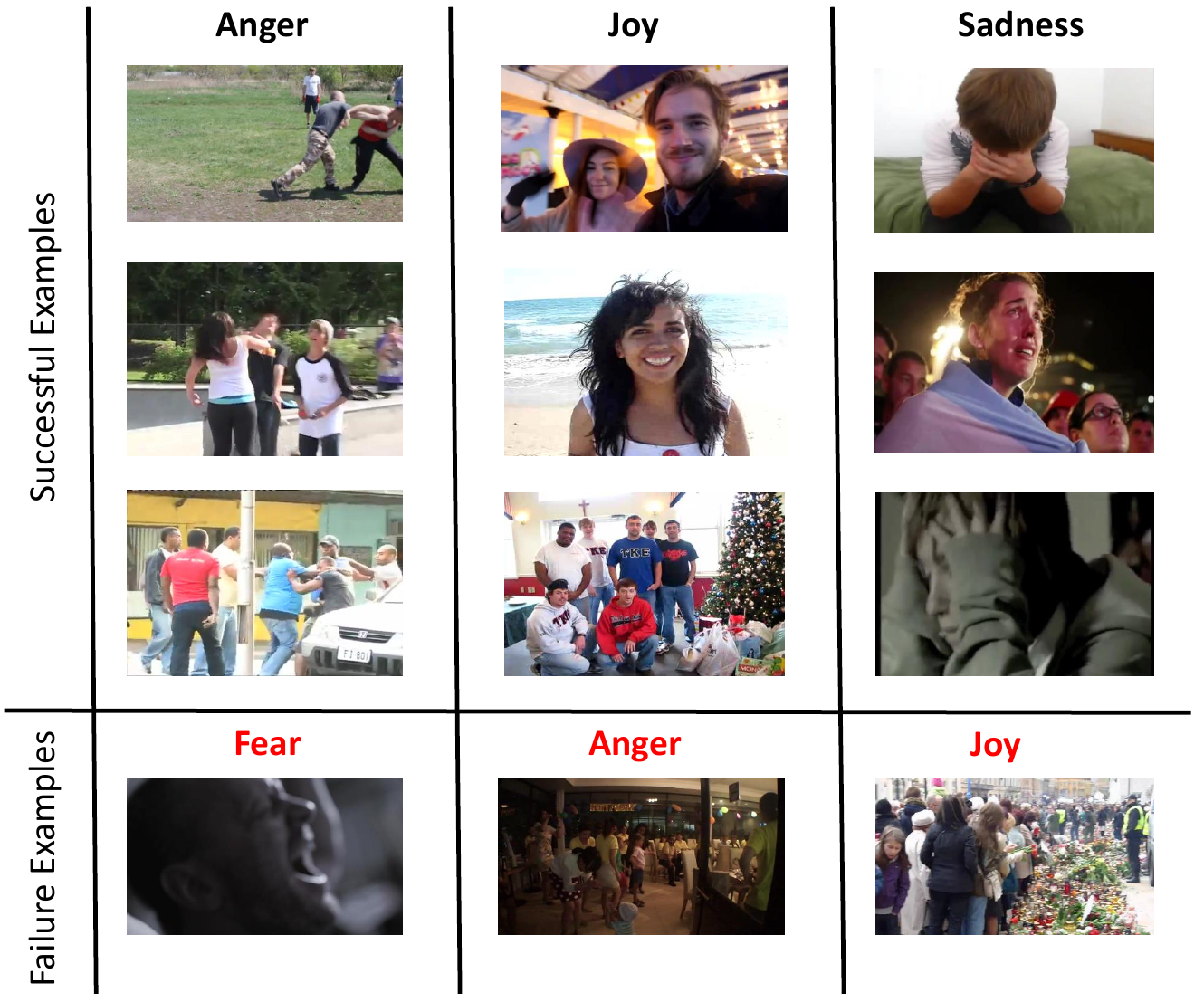} 
\par\end{centering}
\protect\caption{\label{fig:superised-example} Some successful and failure examples
of supervised emotion recognition on the YouTube-8 dataset. The ground
truth categories are given at the top of each column; red labels indicate
classification mistakes. }
\end{figure}

\subsubsection{Hyper-Parameters and Deep Network Configurations}

We conduct further experiments to investigate whether our ITE technique
can maintain its advantage over baselines under different hyper-parameter
settings, CNN configurations, and with additional audio features.
Experimental results show that ITE consistently outperforms\textcolor{red}{{}
}the baselines and suggest ITE's advantage is robust under many different
conditions. For simplicity, our ablation studies cover the Youtube-8
and YF-E6 datasets in supervised setting.

\begin{table}[t]
\centering{}%
\begin{tabular}{|c|c|c|c|c|}
\hline 
Methods & \multicolumn{2}{c|}{ITE} & \multicolumn{2}{c|}{AvgP}\tabularnewline
\hline 
\hline 
Features  & \emph{fc7}  & \emph{fc6}  & \emph{fc7}  & \emph{fc6} \tabularnewline
\hline 
YouTube-8  & 43.8  & \textbf{45.6}  & 41.1  & 42.0 \tabularnewline
\hline 
YF-E6  & \textbf{50.9}  & 49.4  & 48.4  & 48.7 \tabularnewline
\hline 
\end{tabular}\caption{\label{tab:Layer-by-layer-results}Layer-by-layer analysis for supervised
learning. Results obtained from convolutional layers \emph{conv5}
and \emph{conv4} are $22.5\pm2\%$, which are significant lower than
the above. }
\end{table}

\begin{table*}[t]
\begin{centering}
\begin{tabular}{|c|c|c|c|c|c|c|}
\hline 
 & denseSIFT   & MFCC   & ITE($fc7$)   & $\left[\mathrm{ITE}(fc7),{\textstyle \mathrm{denseSIFT}}\right]$
  & $\left[\mathrm{ITE}(fc7),{\textstyle \mathrm{MFCC}}\right]$   & $\left[\mathrm{ITE}(fc7),{\textstyle \mathrm{denseSIFT}},\mathrm{MFCC}\right]$ \tabularnewline
\hline 
\hline 
YouTube-8  & $35.6$   & $44.0$   & $43.8$   & 43.8   & $\mathbf{52.6}$   & 46.7\tabularnewline
\hline 
YF-E6  & $38.6$   & $39.0$   & $50.9$   & 48.8   & \textbf{$\mathbf{51.2}$}{}   & 50.4\tabularnewline
\hline 
\end{tabular}
\par\end{centering}
\textcolor{blue}{\caption{\label{tab:MFCC-results}Concatenated results of hand-crafted feature
and deep features. ITE is computed from $fc7$.}
} 
\end{table*}

 \vspace{0.1in}
\noindent \textbf{Layer-by-layer Analysis.} The CNN we adopt in this
paper, AlexNet, contains 5 convolution layers and 2 fully connected
layers. Although it is generally acknowledged that lower layers preserve
more local information and higher layers contain more global information,
it remains unclear which layers are the most conducive to emotion
detection.
In this experiment, we employ the outputs from the last
and the second last convolutional layers (denoted as \emph{conv5}
and \emph{conv4} respectively) as well as the first and second fully
connected layers (\emph{fc6} and \emph{fc7} respectively) as candidate
features for video frames. We compare the ITE method and the AvgP
method. Table \ref{tab:Layer-by-layer-results} shows the results.

We observe that features from \emph{fc6} perform better for YouTube-8
and features from \emph{fc7} perform only slightly better for YF-E6.
In both cases, the ITE technique outperforms the AvgP technique, suggesting
our encoding mechanism is quite general and is not tied to a particular
layer in the network. Features extracted from the fully connected
layers significantly outperform those from convolutional layers (which
is $22.5\pm2\%$), which suggests that features of higher layers contain
more semantic information that is beneficial for video emotion understanding
task. For the rest of the paper, we use features from \emph{fc7}.

\vspace{0.1in}
\noindent \textbf{Complementarity of CNN and hand-crafted features.}
\quad{}Table \ref{tab:MFCC-results} reports results of 
ITE encoding concatenated with hand-crafted features. We find that (1) results
of concatenation with the visual features (denseSIFT) are comparable
to those of raw ITE on two dataset. This shows that deep features
are not too complementary to visual hand-crafted features. It also
demonstrates the ITE outperforms the traditional hand-crafted features.
(2) The methods that used audio features can achieve very high accuracy
for video emotion recognition. This means that audio track is very
useful for video emotion recognition; (3) The audio hand-crafted features
(MFCC) are very complementary to deep features, since they come from
different `sensors'. (4) Concatenating all features leads to worse
results than that of $\left[ITE(fc7),{\textstyle \mathrm{MFCC}}\right]$
due to the increased dimensionality that resultants from visual hand-crafted
features (and their lack of complementarity).

\vspace{0.1in}
\noindent  \textbf{We test different deep architectures
and find that VGG-16/VGG-19/AlexNet $\mathbf{>}$ GoogLeNet.} \quad{}While
previous experiments showed satisfactory results on emotion analysis
task by using AlexNet architecture, we want to compare with different
architectures to better understand deep features. VGG-16 and VGG-19
\cite{Chatfield14} and GoogLeNet-22 \cite{GoogLeNet2014} achieved
the state-of-the-art for image classification on ImageNet challenge.
Thus we conducted video emotion recognition using high layer features
extracted from these architectures as descriptors. Table \ref{tab:VGG-GoogLeNet-results}
presented the experimental results. We use $fc7$ of $16$ and $19$
layers VGG and $inception-5b$ of GoogLeNet. AvgP is used for all
the deep architectures. The results of VGG-16 and VGG-19 are comparable
to AlexNet, and outperform that of GoogLeNet-22. Although GoogLeNet
gets promising results on image classification task, the lower results
in Table \ref{tab:VGG-GoogLeNet-results} imply that it may not be
the best choice for video emotion recognition.

\vspace{0.1in}
\noindent  \textbf{The number of auxiliary clusters.} The
number of clusters for the auxiliary images (i.e., $D$ in Eq. \ref{eq:similarity})
is a key parameter in our knowledge transfer framework. 
With more clusters, the framework is more capable of capturing
the rich spectrum of emotional information, but is also prone to overfitting.
Here, we empirically test how this parameter affects the ITE performance.
The number of clusters is plotted against supervised performance in
Figure \ref{tab:Varying-class-center-bins}. ITE results gradually
improve when the number of clusters is increased from 100 to 2000.
After 2000, the performance saturates with a slight drop as $D$ keeps
increasing. ITE outperforms AvgP over most settings for $D$, which
indicates that our finding is robust to hyper-parameter settings.
AvgP stays constant since varying $D$
does not affect its performance. 

\begin{table}
\begin{centering}
\begin{tabular}{|c|c|c|c|c|}
\hline 
 & VGG-16   & VGG-19   & GoogLeNet-22   & AlexNet\tabularnewline
\hline 
\hline 
YouTube-8  & $44.7$   & $44.0$   & $35.6$   & $41.1$\tabularnewline
\hline 
YF-E6  & $49.3$   & $48.8$   & $38.3$   & $48.4$\tabularnewline
\hline 
\end{tabular}
\par\end{centering}
\textcolor{blue}{\protect\caption{\label{tab:VGG-GoogLeNet-results}VGG and GoogLeNet results. The AvgP
is used here.}
} 
\end{table}

\noindent 
\begin{figure}[t]
\begin{centering}
\includegraphics[scale=0.2]{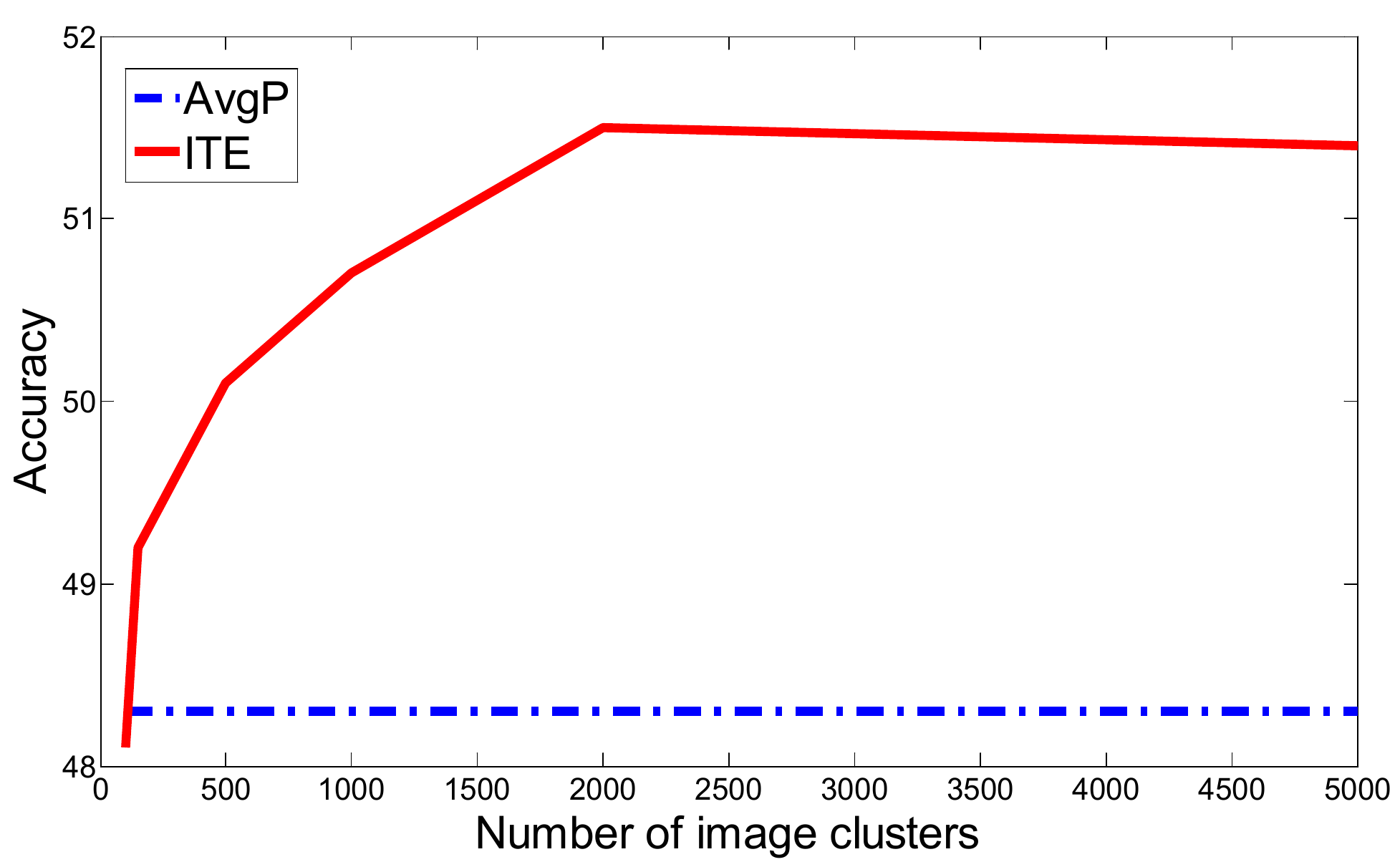} 
\par\end{centering}
\vspace{-0.1in}
 \protect\caption{\label{tab:Varying-class-center-bins}Influence of
varying the number of clusters of auxiliary images for ITE. The X-axis
is varying the number of image clusters of auxiliary data; and y-axis
is the accuracy of recognition tasks. }
\end{figure}

\subsection{Zero-Shot Emotion Recognition}

We conduct zero-shot emotion recognition on YouTube-8, YouTube-24,
and the VideoStory-P14 datasets. The YF-E6 dataset contains only 6
emotion classes. Splitting the 6 further into disjoint training classes
and test classes will lead to difficulties for properly relating unknown
classes to known classes. In the VideoStory-P14 dataset, we use \emph{anger,}
\emph{joy,} \emph{surprise,} and \emph{terror} as testing classes,
with a total of $300$ testing instances. For YouTube-8, we use \emph{fear}
and \emph{sadness} as the testing classes. For YouTube-24, we randomly
split the 24 classes into $18$ training and $6$ testing classes
with 5-round repeated experiments. In the zero-shot setting, no instances
in test classes are seen during training.

We compare our T1S algorithm with Direct Attribution Prediction (DAP)
\cite{lampert13AwAPAMI,lampert2009zeroshot_dat}. For DAP, at test
time each dimension of the word vectors of each test sample is predicted,
from which the test class labels are inferred. DAP can be understood
as directly using Eq (\ref{eq:self-training-1-step}) without the
word vector smoothing by Eq (\ref{eq:self-training}). Four variants
are compared: (a) using different video-level feature representation
(AvgP or ITE); (b) using different zero-shot learning algorithm (T1S
or DAP).

Figure \ref{tab:Zero-shot-Learning-results} shows the results. Our
ITE+T1S approach produces the best accuracy, outperforming the second
best baseline by 3.6, 4.8, and 1.2 absolute percentage points respectively
and the random baseline by 8.1, 6.3, and 15.9 absolute percentage
points. We observe that AvgP+T1S is the second best technique on VideoStory-P14
and YouTube-24, but ITE+DAP is the second best technique on the YouTube-8
dataset. An important difference between the two scenarios is that
YouTube-8 contain less emotions than VideoStory-P14 and YouTube-24,
so the semantic distance between individual emotions is greater in
YouTube-8. This suggests the T1S technique contributes the biggest
performance gain when the training classes bear some similarity to
the unseen test classes. However, when the training classes are very
different from the testing classes, the ITE encoding scheme plays
an important role. It is also worth mentioning that the results of
YouTube-24 have a largest margin improvement over baselines than the
two other datasets. This result indicates zero-shot learning performs
better when a larger variant set of emotions exist in the training
set. Overall, the experiments show the combination of ITE+T1S is effective
under different zero-shot learning conditions. Given the inherent
difficulties of the zero-shot learning task, we consider the results
to be very promising.

\vspace{0.1in}
\noindent  \textbf{Qualitative results.} In Figure \ref{fig:ZSL-example-Y24},
we show some successful examples of zero-shot emotion prediction.
We highlight that even without any training examples on these categories,
our method can still classify these video successfully using the encoded
feature. Thus considering the difficulty of zero-shot emotion prediction,
our results are very promising.

Note that Ekman dataset is not used for this tasks
due to the small number of emotion classes. Specifically, in our work,
each class-level emotion textual name $w\in\mathcal{V}$ is projected
into a $\mathcal{K}$-dimensional embedding vector $\pmb\psi_{w}\in\mathbb{R}^{\mathcal{K}}$
in the semantic word vector space; a regressor function $g\left(\cdot\right)$
is trained from video-level features to the corresponding embedding
vector $\pmb\psi_{w}$. In zero-shot learning scenarios, we need to
further split the 6 emotion classes of Ekman dataset into auxiliary
and testing dataset. In other words, we only have at most 4 embedding
vectors $\pmb\psi_{w}$ to train the regressor $g\left(\cdot\right)$
(in the split of 4 auxiliary and 2 testing classes). It is however
extremely hard to train a reasonable regressor (without overfitting)
with only 4 embedding vectors. 


\begin{figure}[t]
\begin{centering}
\includegraphics[scale=0.26]{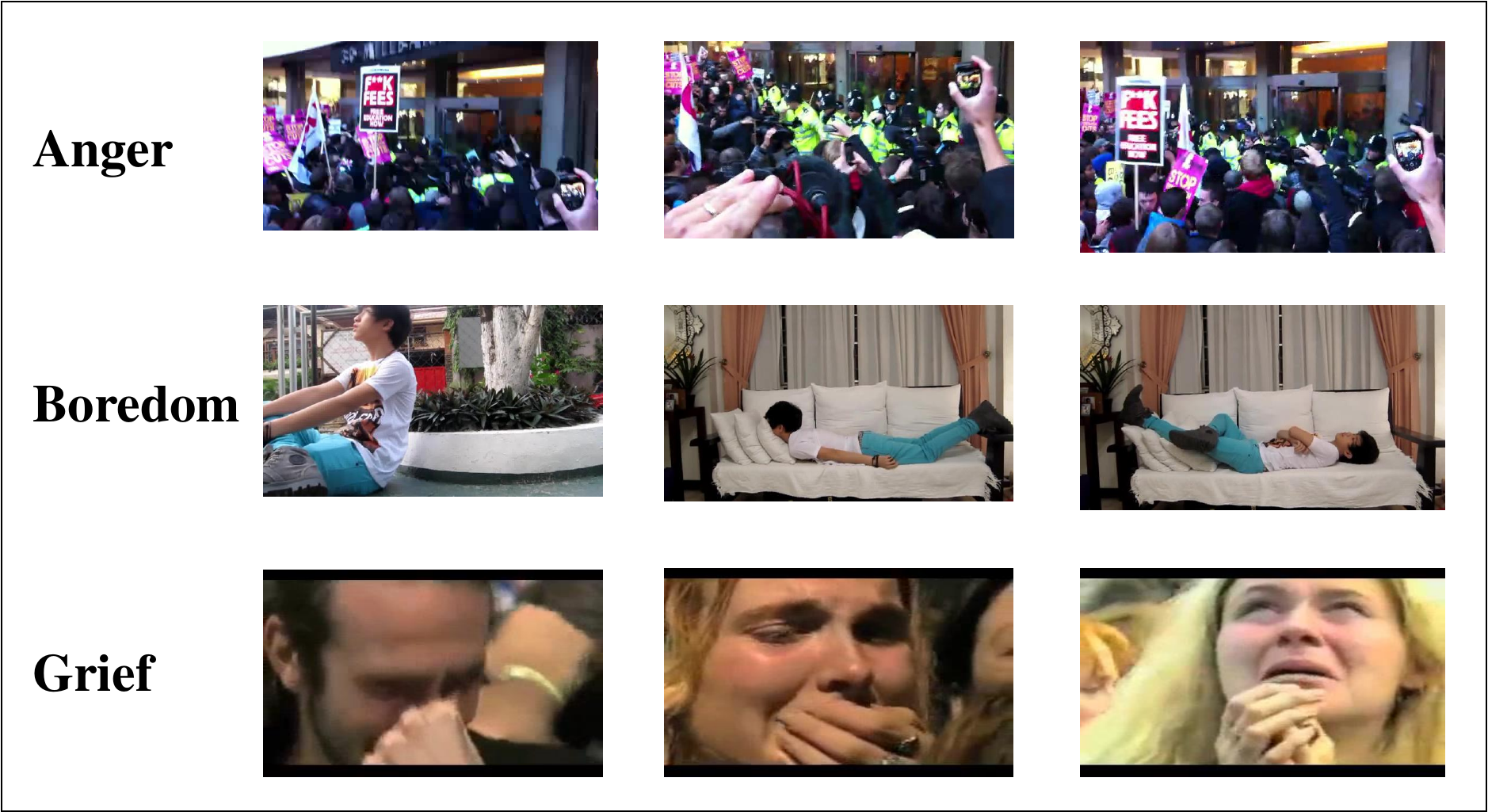} 
\par\end{centering}
\protect\protect\protect\caption{\label{fig:ZSL-example-Y24} Qualitative results of zero-shot emotion
recognition. We show the keyframes of three successful cases: the
frames of top row shows a video clip of an anger parade; the middle
row is about a video of a boredom boy walking and lying on the couch;
The bottom row is for the grief reaction of fans when their favorite
football team lose the game.}
\end{figure}

\begin{figure}[t]
\begin{centering}
\includegraphics[scale=0.21]{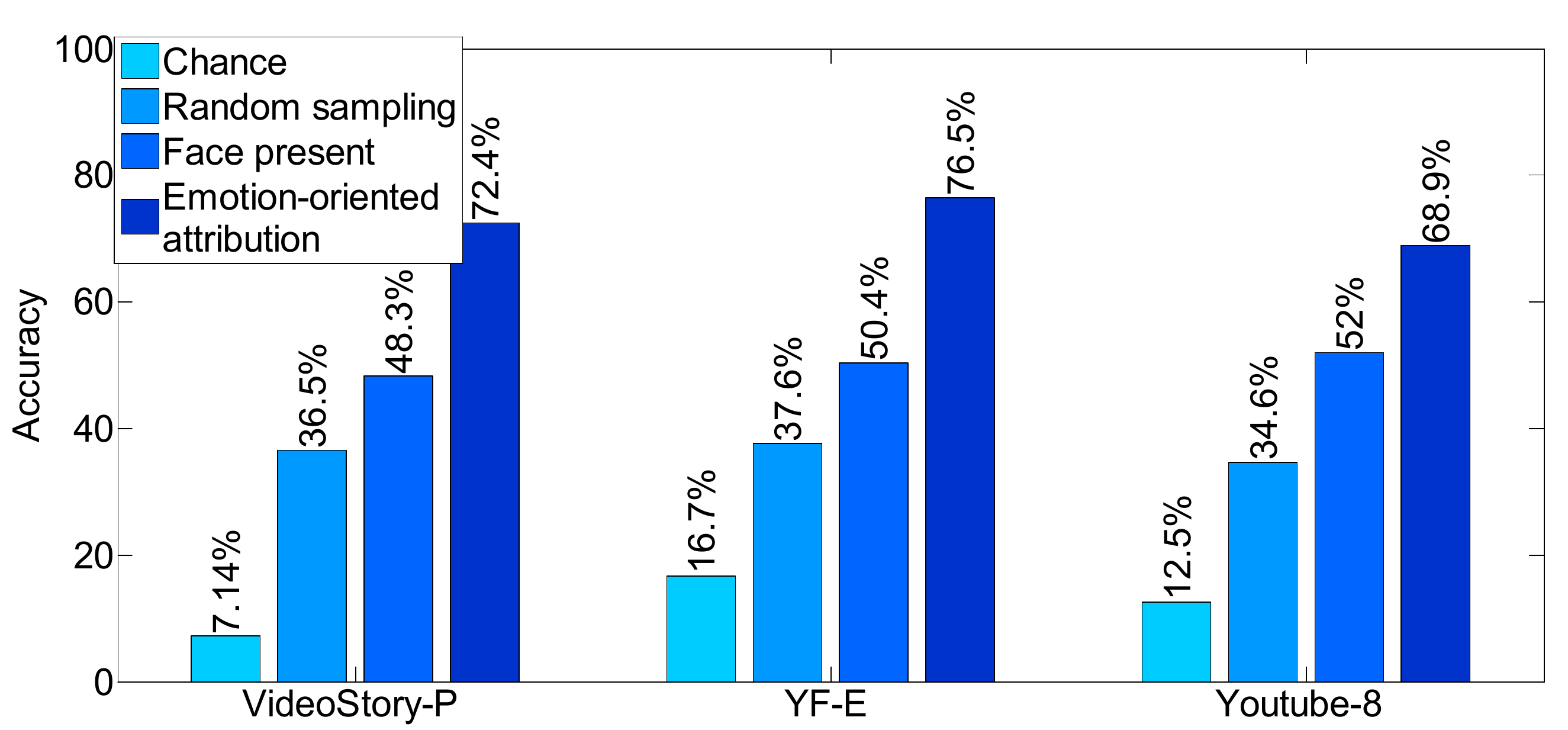} 
\par\end{centering}
\protect\protect\protect\caption{\label{tab:User-guess-evaluating} Quantitative evaluation of video
emotion attribution using the YouTube-8 dataset.}
\end{figure}

\begin{table}[t]
\centering{}\centering \protect\caption{Reliability evaluation of video emotion attribution experiments. We
random split 10 annotators into 2 groups with 5 person each and compute
the each group's score. The accuracy of each method
is reported here.}
\label{tab:User-guess-measurement} %
\begin{tabular}{|c|c|c|c|}
\hline 
 & \multicolumn{3}{c|}{Group-1}\tabularnewline
\hline 
 & Random   & Face\_present   & Emotion \tabularnewline
\hline 
VideoStory-P14   & 35.1   & 41.7   & \textbf{65.9}\tabularnewline
\hline 
YF-E6   & 34.4   & 54.1   & \textbf{71.3}\tabularnewline
\hline 
YouTube   & 32.5   & 45.5   & \textbf{62.8}\tabularnewline
\hline 
 & \multicolumn{3}{c|}{Group-2}\tabularnewline
\hline 
 & Random   & Face\_present   & Emotion \tabularnewline
\hline 
VideoStory-P14   & 37.9   & 54.9   & \textbf{78.9}\tabularnewline
\hline 
YF-E6   & 40.8   & 46.7   & \textbf{81.7}\tabularnewline
\hline 
YouTube   & 36.7   & 58.5   & \textbf{75.0}\tabularnewline
\hline 
\end{tabular}
\end{table}

\begin{figure}[t]
\begin{centering}
\includegraphics[scale=0.28]{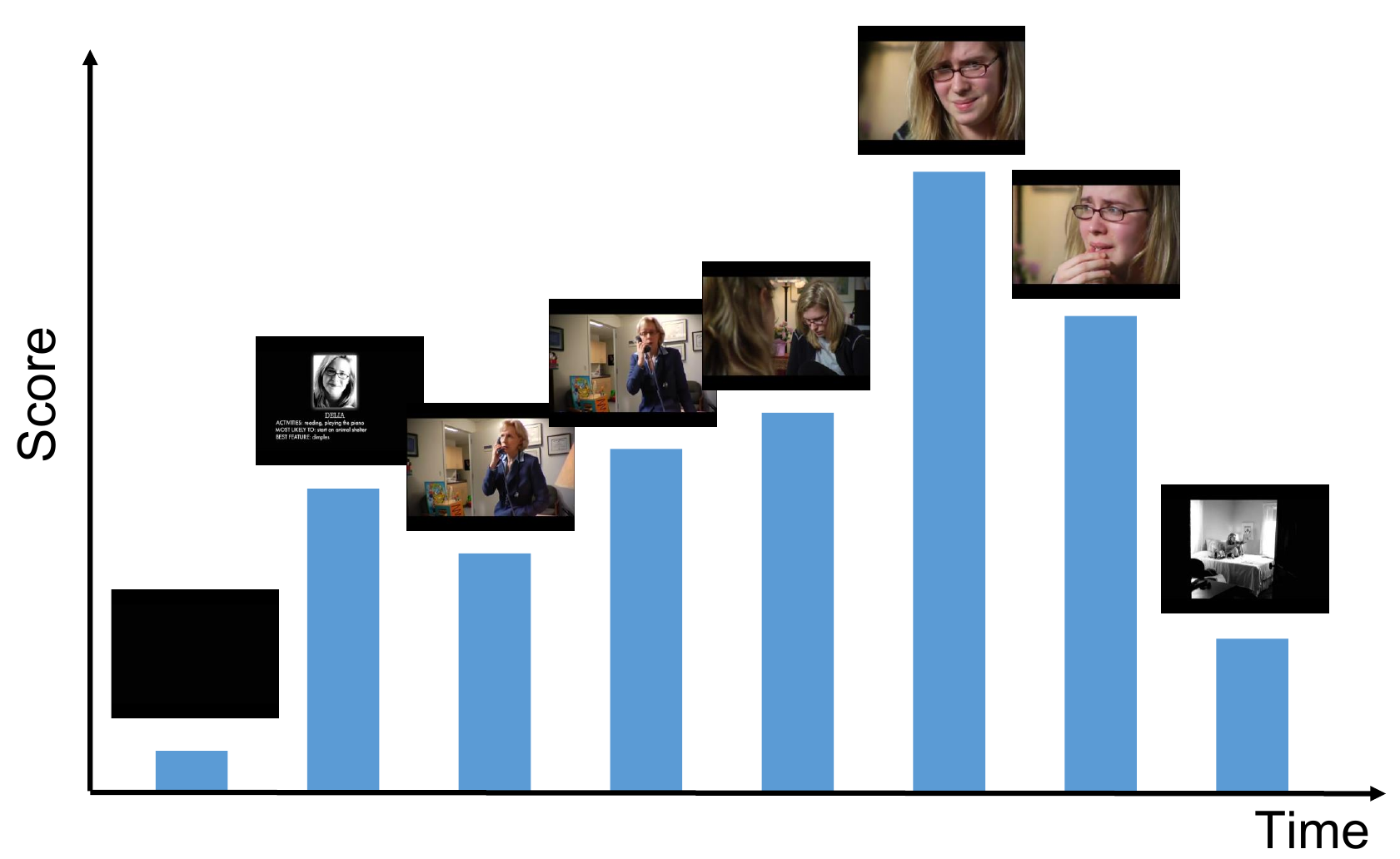} 
\par\end{centering}
\vspace{-0.1in}
 \protect\protect\protect\caption{\label{tab:User-guess-evaluating2} An example of video emotion attribution.}
\end{figure}

\subsection{Video Emotion Attribution}

As discussed earlier, another advantage of our encoding scheme is
that we can identify the video clips that have high impact on the
overall video emotion. A pilot study we performed indicated that emotions
are sparsely expressed in videos. On average, around $10\%$ of 
video frames are related to emotion in our three datasets.

As the first work on video emotion attribution, we define the evaluation
protocol of user study to evaluate the performance of different algorithms
for this task: Ten participants, unaware of project goals, were invited
for the user study. Given all emotion keywords of the corresponding
dataset and clip computed from the video, participants are asked to
guess the name of the emotion expressed in the clip. These clips are
generated by different baseline techniques, as discussed later. Since
the ground-truth video emotion labels are known, we computed the fraction
of participants who assigned the correct emotion label for each clip.

We randomly select $20$ videos from each of the three datasets. For
each video, we extract a 2-second video clip that contains the highest
attribution towards video emotion, using Eq (\ref{eq:maximum_attribution}).

For comparison purposes, we created the following baselines: \textbf{Chance},
which is the probability of correctly guessing the emotion. \textbf{Random
sampling}, where we first randomly sample 2 non-overlapping clips
of $2$ seconds each from the same video. We use both clips in the experiments
and compute the average score as the results of this method. \textbf{Face
presence}, where we use the ``face\_present'' feature \cite{face_detection}
to rank all the videos frames; frames with larger and more faces are
ranked higher. One clip of 2-second length is generated for each video
by using the top ranked frames.

The results are shown in Fig.~\ref{tab:User-guess-evaluating}. Our
method achieve best accuracy, and outperforms the Face\_present baseline
by 16-26 absolute percentage points. Although the presence of a human
face is often correlated with the expression of emotions, many user-generated
videos in our datasets express emotions through other channels like
body language or color. Thus, our technique compared more favorably
to the face presence baseline. These results indicate that our method
can consistently identify video clips that convey emotions recognizably
similar to the emotion conveyed by the original video.

To further validate the reliability of the attribution experiments
in this user study, we randomly split 10 participants into 2 groups
and measure each group's score in Table \ref{tab:User-guess-measurement}.
The measurement shows the consistency of our result; and in each group,
our experimental conclusion still hold: the ITE result achieves best
accuracy between two groups. .

A qualitative result of emotion attribution is shown in Figure~\ref{tab:User-guess-evaluating2},
where the video is uniformly sampled every 10 frames. The bar chart
shows scores of different frames, where the key frames are shown above
the bars. The figure demonstrates that clips with stronger emotional
contents are given higher scores of attribution, validating the effectiveness
of our method.

\begin{figure}
\begin{centering}
\includegraphics[scale=0.22]{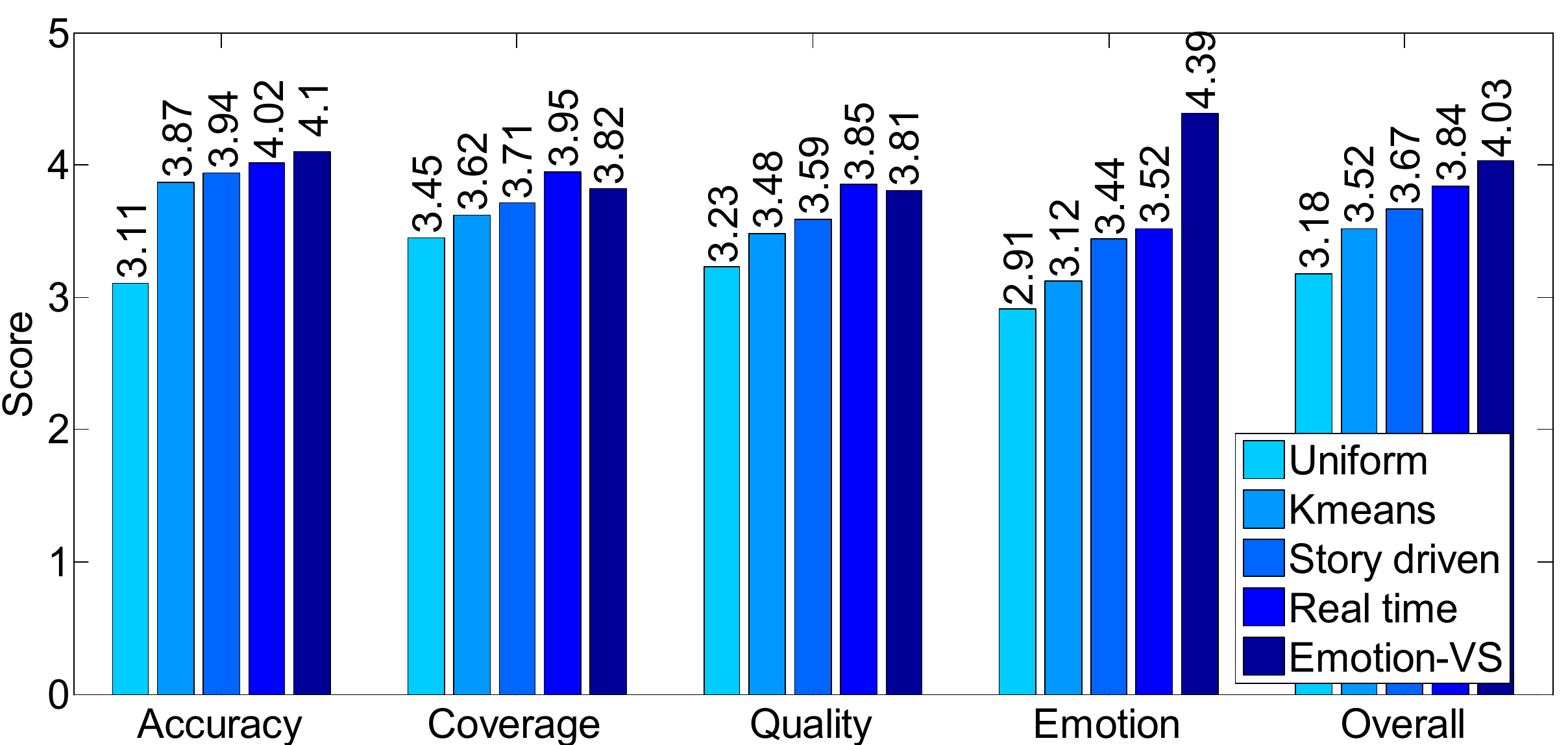} 
\par\end{centering}
\protect\protect\protect\caption{\label{tab:The-user-study-of}User-study results on the video summarization.}
\end{figure}

\begin{figure*}[thb]
\begin{centering}
\includegraphics[scale=0.41]{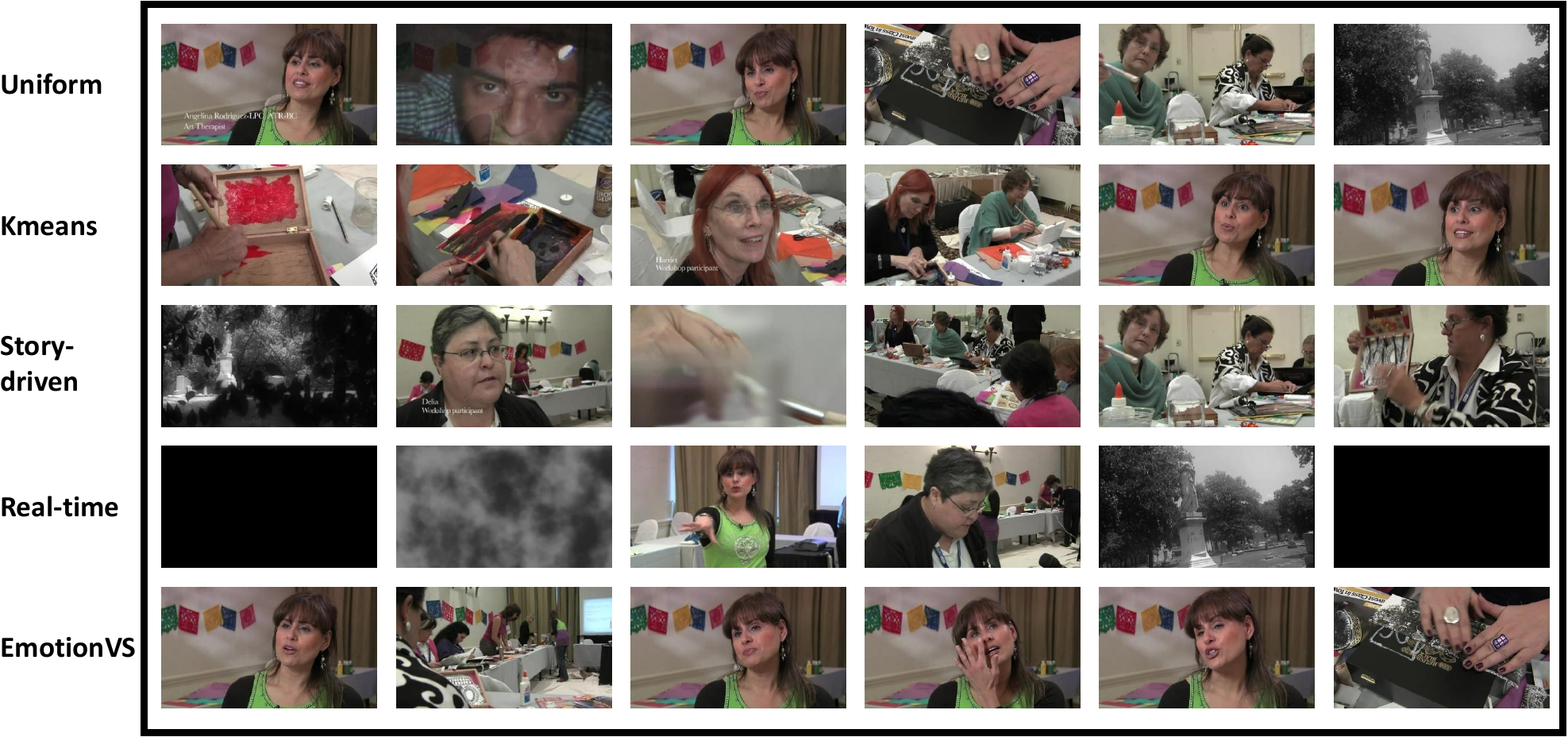} \\
 \includegraphics[scale=0.41]{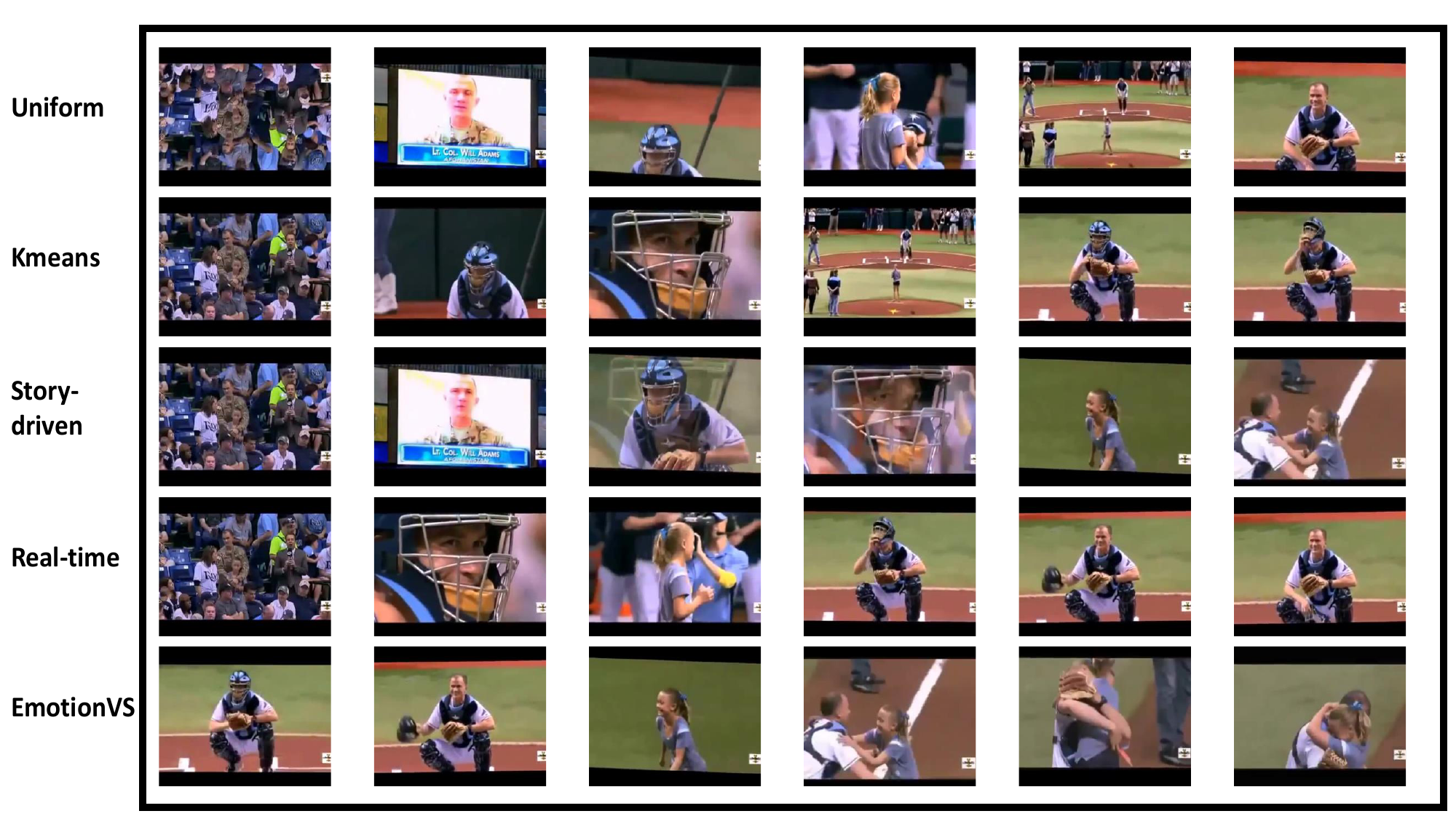} 
\par\end{centering}
\begin{centering}
\vspace{-0.1in}
\par\end{centering}
\protect\caption{\label{fig:summary_results}The qualitative results of emotion-oriented
video summarization. }
\end{figure*}

\subsection{Emotion-Oriented Video Summarization}

\begin{figure}
\begin{centering}
\includegraphics[scale=0.32]{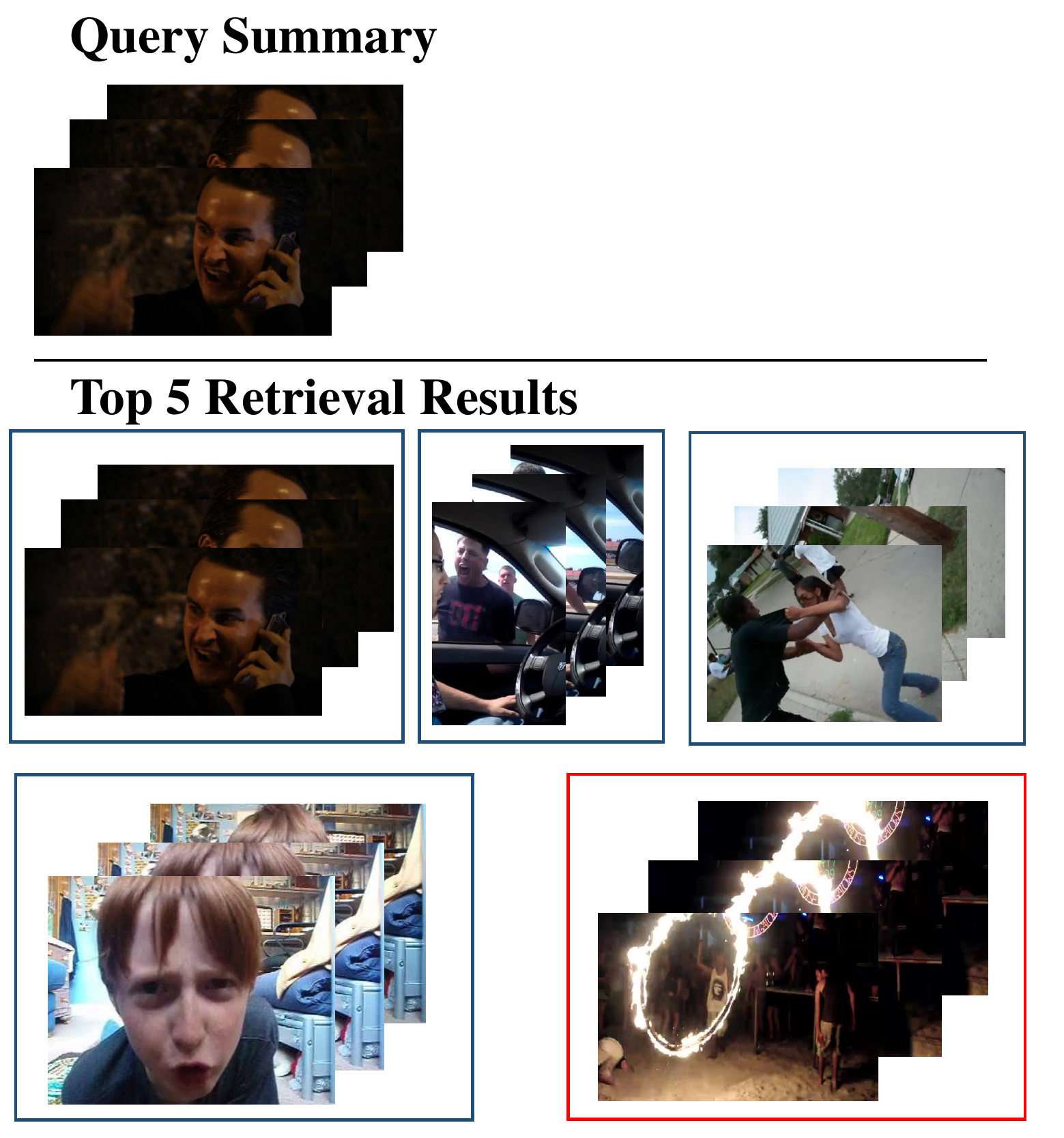} 
\par\end{centering}
\caption{\label{fig:Retrieval_result} A retrieval example using video summary.
The query summary of an angry man is illustrated on the top. Top 5
retrieval results are showed on the bottom. The first 4 results (blue
box) are all in the 'angry' category and the last result (red box)
is a clip of magic show from 'joy' category.}
\end{figure}

Finally, we evaluate our framework on emotion-oriented video summarization.
We compare with four baselines: (1)~\textbf{Uniform sampling}, which
uniformly samples several clips from video. (2)~\textbf{K-means sampling},
which simply clusters the clips and selects a clip closest to each
cluster centroid. (3) \textbf{Story-driven summarization~\cite{story-driven-summary}}.
This approach was developed to summarize very long egocentric videos.
We slightly modify the implementation and make the length of the summary
controllable for our task. (4) \textbf{Real-time summarization~\cite{DBLP:conf/mm/WangJCGDW14}},
which is a technique aimed at efficient summarization of videos based
on semantic content recognition results. For all the methods, the
length of summary is fixed to 6 seconds if the original video is longer
than 1 minute. For short videos, the length is fixed to $10\%$ of
the original video.

Following \cite{Truong:2007:VAS:1198302.1198305}, we conduct a user
study to evaluate different summarization methods. Ten subjects unfamiliar
with the project participated in the study. We show the summary results
of all the methods (without the audio information) to each participant.
Participants are asked to rate each result on a five-point scale for
each of the following evaluation metrics: (1) \emph{Accuracy}: the
summary accurately describes the ``dominating high-level semantics
of the original video"; (2) \emph{Coverage}: the summary covers as
much visual content using as few frames as possible. (3) \emph{Quality}:
the overall subjective quality of the summary; (4) \emph{Emotion}:
the summary conveys the same main emotion as the original video.

The results are shown in Figure \ref{tab:The-user-study-of}. The
average score is shown in the ``Overall'' column. Our method (``Emotion-VS'')
performs better than the other methods on the accuracy and the emotion
metrics. On the emotion metric, we beat the best baseline by a margin
of 0.87. Although we are doing slightly worse on the coverage metric
(-0.13 compared to the best baseline), the drop in quality is minimal
(-0.04 compared to the best baseline). The results suggest that the
selection of emotional key frames and clips does not only capture
the emotion of the original video, but also improves the overall accuracy
of the summary, since emotional content plays an important role in
an accurate summary. Our emotion-oriented summarization method significantly
increases the amount of emotional contents captured by the summary
without material loss on other quality measures.

We show a qualitative evaluation in Figure \ref{fig:summary_results}.
At the top, the figure shows a video of an art therapist (the woman
in green). Different from other methods, our summarization not only
captured the therapy procedure, but also focused on the sadness of
the therapist, which is the central emotion conveyed in this video.
At the bottom, we illustrate a user-generated video where a father
surprises his daughter during a baseball game by dressing as the catcher
and revealing himself. All baseline methods are more focused on the
baseball game itself, which is only marginally related to the
emotion of this video. In contrast, our method clearly captures the
reveal of the father, the surprised daughter, and the subsequent
emotional hug.

Figure \ref{fig:Retrieval_result} demonstrates an example of using
the video summary for retrieval on all available videos in terms of
cosine similarity between videos and frames in Eq (\ref{eq:maximum_attribution}).
The results shows the top retrieval results are all the same category
and share some common visual characteristics. This also indicates
the effectiveness of our method when finding the emotional clips.

\section{Conclusions}

Making a strong emotional appeal to viewers is the ultimate goal of
many video producers. Therefore, being able to recognize a video's
emotional impact is an important task for computer vision and affective
computing. Given the diverse landscape of emotional expressions, we
propose the first knowledge transfer framework for learning from heterogeneous
sources for the task of understanding video emotion. Within the framework,
we tackled inter-related video understanding problems including supervised
and zero-shot emotion recognition, emotion attribution and emotion-oriented
summarization.

For effective knowledge transfer, we learn encoding schemes from a
large-scale emotional image data set and a large, 7-billion-word text
corpora. This transfer facilitates the creation of a representation
conducive to the tasks of understanding video emotion. In zero-shot
emotion recognition, an unknown emotional word is related to known
emotion classes through the use of a distributed representation in
order to identify emotions unseen during training. Our experiments
on three challenging datasets clearly demonstrate the benefit of utilizing
external knowledge. Our framework also enables novel applications
such as emotion attribution and emotion-oriented video summarization.
A user study shows that our summaries accurately capture emotional
content consistent with the overall emotion of the original video.

As the future work, we will address the joint application of emotion-oriented
summarization and story-driven summarization, which should allow us
to create complete and emotionally compelling stories.   
We will also study the scheme of encoding of motion  in the future. Currently,
we do not have the large scale of auxiliary emotional ``motion'' dataset which can facilitate our whole framework. Such an auxiliary dataset, however is essential for the knowledge transfer to recognize video emotion in our tasks.

\section{Acknowledgement}

We are grateful to the anonymous reviewers, whose suggestions considerably
improved this paper. This work was supported in part by a National
863 Program ($\#2014AA015101$) and a grant from the NSF China ($\#61572134$).

\bibliographystyle{abbrv}
\bibliography{video_senti_v2}

\newpage  \begin{IEEEbiography} [{\includegraphics[width=1in,height=1.25in,clip,keepaspectratio]{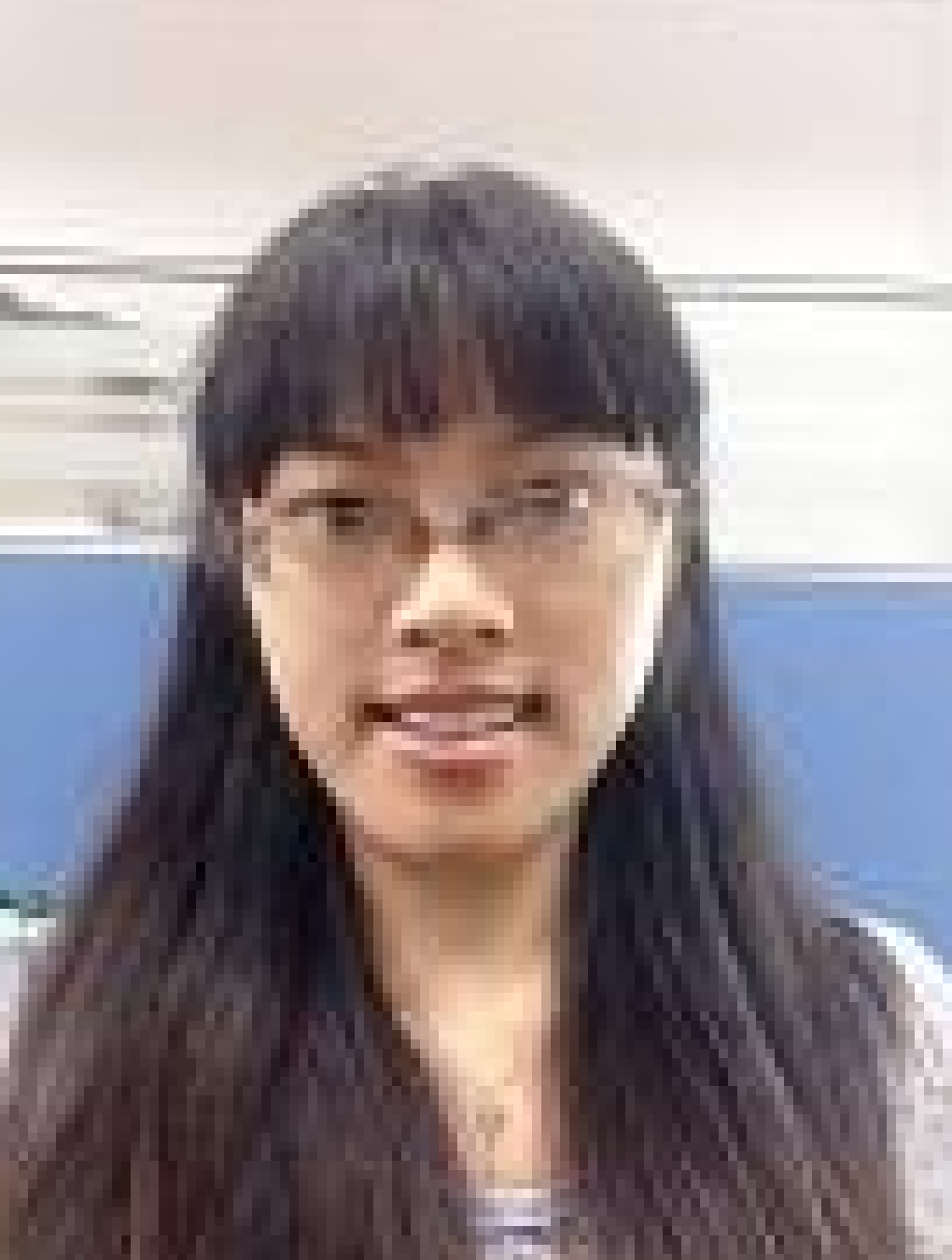}}]{Baohan Xu} received  the BS degree from Fudan University, Shanghai, China, in 2014. She is now pursuing her MS degree of Computer Science at Fudan University. Her research interests include computer vision and video emotion analysis. \end{IEEEbiography}

\vspace{-1cm} 
\begin{IEEEbiography}[{\includegraphics[width=1in,height=1.3in,clip,keepaspectratio]{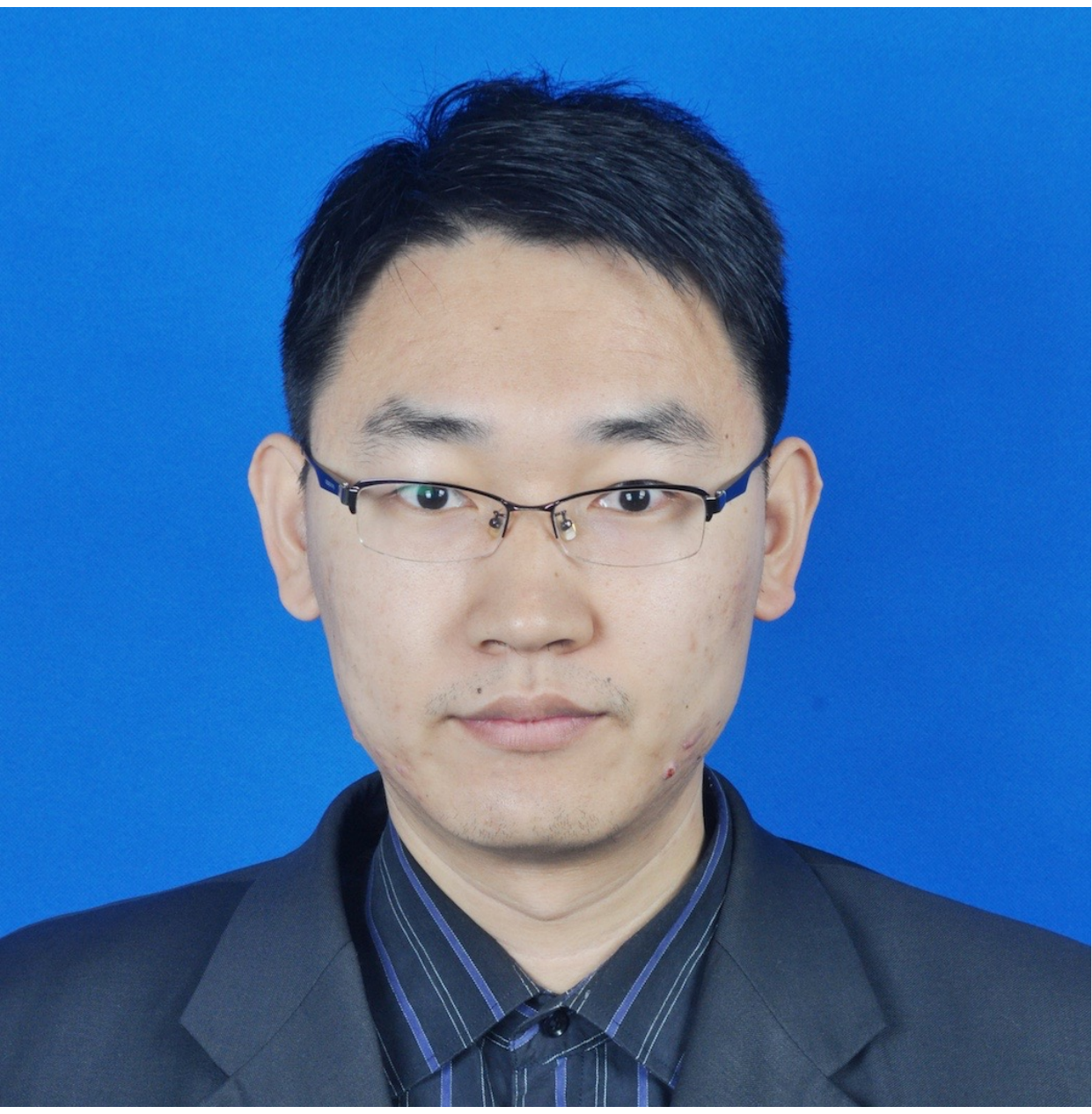}}]{Yanwei Fu} received the PhD degree from Queen Mary University of London in 2014, and the MEng degree in the Department of Computer Science \& Technology at Nanjing University in 2011, China. He worked as a Post-doc in Disney Research at Pittsburgh from 2015-2016. He is currently an Assistant Professor at Fudan University. His research interest is image and video understanding, and life-long learning. \end{IEEEbiography}
\vspace{-1cm} 
\begin{IEEEbiography} [{\includegraphics[width=1in,height=1.25in,clip,keepaspectratio]{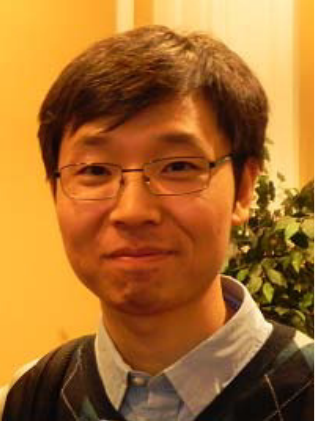}}] {Yu-Gang Jiang} is a Professor in School of Computer Science, Fudan University, China. His Lab for Big Video Data Analytics conducts research on all aspects of extracting high-level information from big video data, such as video event recognition, object/scene recognition and large-scale visual search. He is the lead architect of a few best-performing video analytic systems in worldwide competitions such as the annual U.S. NIST TRECVID evaluation. His visual concept detector library (VIREO-374) and video datasets (e.g., CCV and FCVID) are widely used resources in the research community. His work has led to many awards, including "emerging leader in multimedia" award from IBM T.J. Watson Research in 2009, early career faculty award from Intel and China Computer Federation in 2013, the 2014 ACM China Rising Star Award, and the 2015 ACM SIGMM Rising Star Award. He holds a PhD in Computer Science from City University of Hong Kong and spent three years working at Columbia University before joining Fudan in 2011. 
\end{IEEEbiography}

\vspace{-1cm} 
\begin{IEEEbiography}[{\includegraphics[width=1in,height=1.25in,clip,keepaspectratio]{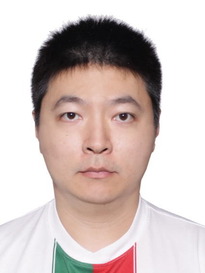}}]{Boyang Li} is a Research Scientist at Disney Research, where he directs the Narrative Intelligence group. He obtained his Ph.D. in Computer Science from Georgia Institute of Technology in 2014, and his B. Eng. from Nanyang Technological University, Singapore in 2008. His research interests include computational narrative intelligence, or the creation of Artificial Intelligence that can understand, craft, tell, direct, and respond appropriately to narratives, and understanding how human cognition comprehends narratives and produces narrative-related affects. 
\end{IEEEbiography}
\vspace{-1cm} 
\begin{IEEEbiography}[{\includegraphics[width=1in,height=1.25in,clip,keepaspectratio]{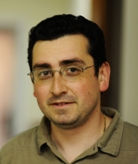}}]{Leonid Sigal}  is a Senior Research Scientist at Disney Research Pittsburgh and an adjunct faculty at Carnegie Mellon University. Prior to this he was a postdoctoral fellow in the Department of Computer Science at University of Toronto. He completed his Ph.D. at Brown University in 2008; he received his B.Sc. degrees in Computer Science and Mathematics from Boston University (1999), his M.A. from Boston University (1999), and his M.S. from Brown University (2003). From 1999 to 2001, he worked as a senior vision engineer at Cognex Corporation, where he developed industrial vision applications for pattern analysis and verification.  
\end{IEEEbiography}
\end{document}